\documentclass{article}
\usepackage{ijcai26}

\PassOptionsToPackage{table, dvipsnames}{xcolor}

\usepackage{times}
\usepackage{soul}
\usepackage{url}
\usepackage[hidelinks]{hyperref}
\usepackage[utf8]{inputenc}
\usepackage[small]{caption}
\usepackage{graphicx}
\usepackage{amsmath}
\usepackage{amsthm}
\usepackage{natbib}
\usepackage{booktabs}
\usepackage{algorithm}
\usepackage{algorithmic}
\usepackage[switch]{lineno}

\usepackage{xcolor}      % Handles colors and [table] via PassOptions
\usepackage{subcaption}  % For subfigures (Figure 1a, 1b)
\usepackage{pifont}      % For special symbols like \ding
\usepackage{multirow}    % For merging rows in tables
\usepackage{makecell}    % For line breaks inside table cells
\usepackage{tabularx}    % For flexible width tables
\usepackage{colortbl}    % Extra table coloring control
\usepackage{array}       % For custom column types
\usepackage{float}       % Better control over figure placement

\definecolor{customblue}{HTML}{33BBEE}   
\definecolor{customorange}{HTML}{EE7733}
\definecolor{custommagneta}{HTML}{EE3377}
\definecolor{customolive}{HTML}{999933}
\definecolor{cusomgray}{HTML}{BBBBBB}
\newcommand{\greencircle}{{\textcolor{green}{\scalebox{0.6}{\ding{108}}}}}
\newcommand{\bluelozenge}{{\textcolor{customblue}{\scalebox{0.6}{\ding{108}}}}}
\newcommand{\orangelozenge}{{\textcolor{customorange}{\scalebox{0.6}{\ding{108}}}}}

\newcommand{\up}[1]{\textcolor{OliveGreen}{\ensuremath{#1}}}
\newcommand{\down}[1]{\textcolor{Red}{#1}}

\title{CardioBench: Do Echocardiography Foundation Models Generalize Beyond the Lab?}% \author{
\author{
Darya Taratynova$^1$\thanks{Equal contribution.}
\and
Ahmed Aly$^1$\footnotemark[\value{footnote}]\and
Numan Saeed$^1$\thanks{Joint supervision.}\And
Mohammad Yaqub$^1$\footnotemark[\value{footnote}]\\
\affiliations
$^1$ Mohamed bin Zayed University of Artificial Intelligence (MBZUAI), Abu Dhabi, UAE\\
\emails
\{darya.taratynova, ahmed.aly, numan.saeed, mohammad.yaqub\}@mbzuai.ac.ae
}
\begin{document}
\maketitle

\begin{abstract}
Foundation models are reshaping medical imaging, yet their application in echocardiography remains limited, hindered by a heavy reliance on private datasets that prevent reproducible comparison. Echocardiography poses unique challenges, including noisy acquisitions, high frame redundancy, and limited diverse public datasets. To address this, we introduce CardioBench\footnote{Code: \url{https://github.com/BioMedIA-MBZUAI/CardioBench}}\footnote{Supplementary Material: \url{https://arxiv.org/abs/2510.00520}}, a comprehensive benchmark for echocardiography foundation models. Specifically, CardioBench unifies eight publicly available datasets into a standardized suite spanning four regression and five classification tasks, covering functional, structural, diagnostic, and view recognition endpoints. Leveraging this framework, we evaluate several leading foundation models, including cardiac-specific, biomedical, and general-purpose encoders, under consistent zero-shot, probing, and alignment protocols. Our analysis reveals that while general-purpose encoders transfer well and often close the gap with probing, they struggle significantly with fine-grained distinctions like view classification and subtle pathology recognition. Results indicate that models capturing temporal cardiac dynamics perform best on functional tasks, while retrieval-based approaches generalize more consistently across datasets. By releasing preprocessing, splits, and public evaluation pipelines, CardioBench establishes a reproducible reference point to guide the architectural design of future echocardiography and possibly other medical imaging foundation models.
\end{abstract}

\begin{figure}[t]
    \centering
    \includegraphics[width=0.7\linewidth]{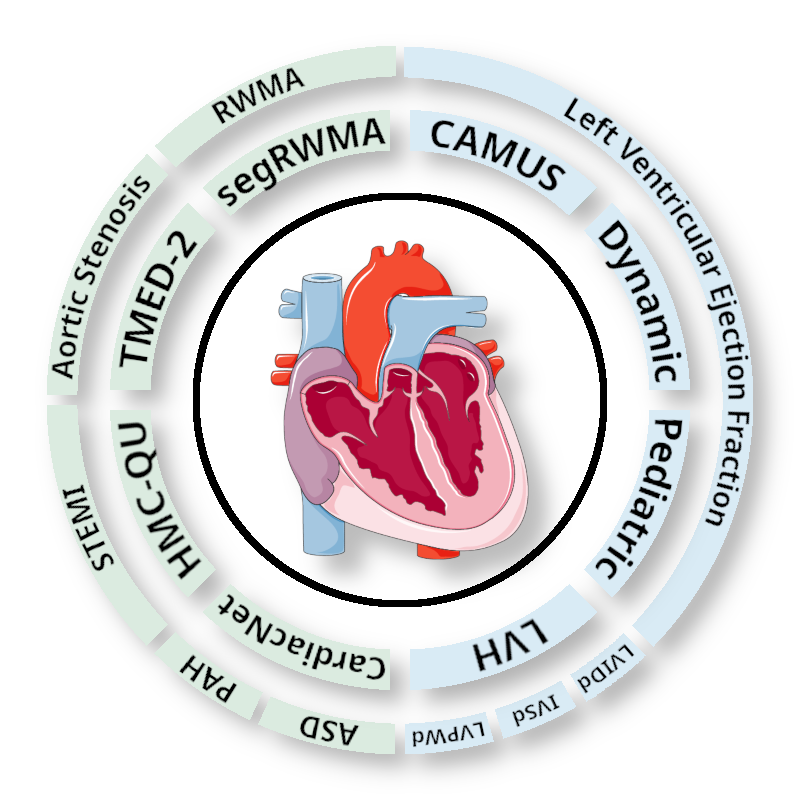} 
    \caption{CardioBench: A unified benchmark for echocardiography.}
    \label{fig:heart}
\end{figure}
\section{Introduction}

Foundation models have become a transformative force in vision and language domains, demonstrating remarkable zero-shot generalization across diverse tasks including image classification, retrieval, visual grounding, and multimodal reasoning (\cite{jia2021scaling, ghiasi2022scaling,li2022grounded, singh2022flava,alayrac2022flamingo}). Large-scale architectures such as CLIP, DINOv3, and SigLIP2 demonstrate that self-supervised and multimodal learning produce general-purpose backbones with strong transferability across downstream tasks (\cite{radford2021learning,simeoni2025dinov3,tschannen2025siglip}). This success has extended to medical imaging, where foundation models have advanced disease classification in chest radiography (\cite{irvin2019chexpert,johnson2019mimic}) and achieved state-of-the-art segmentation in volumetric CT and MRI (\cite{roy2023mednext,huang2023stu}). These advances have been enabled by the availability of large, standardized public datasets that allow for reproducible benchmarking and fair model comparison. 

\begin{figure*}[t]
        \centering
        \includegraphics[width=\linewidth]{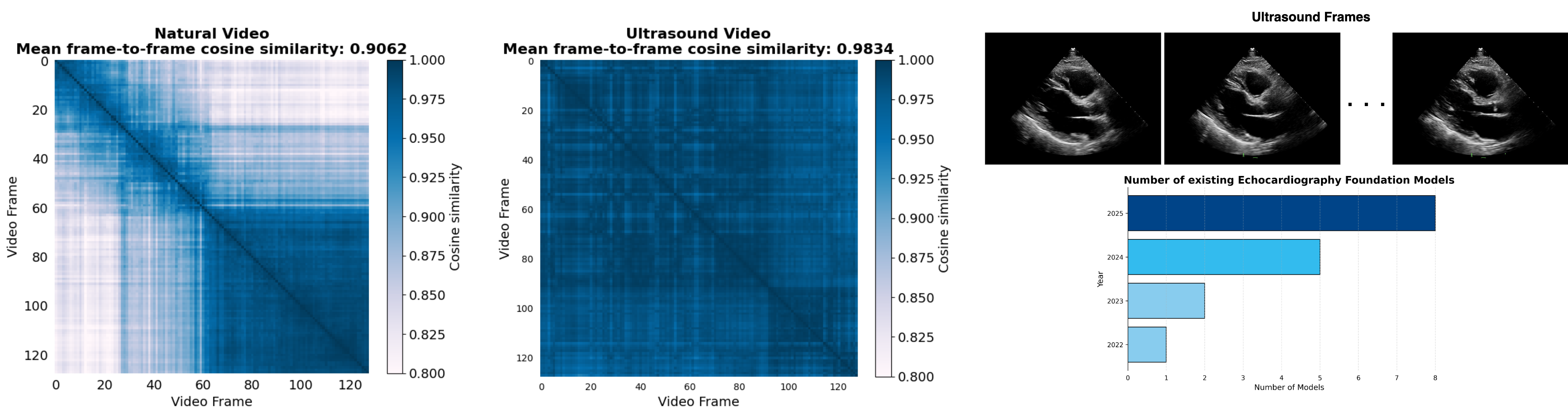}
        \caption{The figure on the left shows frame-level cosine similarity matrices: natural video frames from the SumMe dataset (\cite{gygli2014creating}) versus echocardiography video frames extracted using SigLIP2 (\cite{tschannen2025siglip}). Echocardiography videos exhibit much higher frame-to-frame similarity compared to natural videos, making informative feature extraction more challenging. The figure on the right shows the number of echocardiography foundation models released each year.}
        \label{fig:general_fig1}

\end{figure*}

In contrast, echocardiography remains underserved despite its central role in clinical practice. It is a first-line, non-invasive imaging modality for assessing cardiac structure and function, and is integral to routine cardiovascular diagnosis and management (\cite{mitchell2019guidelines}). It is used to evaluate a wide range of cardiac conditions from multiple views, necessitating models that can generalize across overlapping tasks rather than addressing each in isolation. Therefore, there is growing interest in developing ultrasound foundation models, as evidenced by the increasing number of models proposed each year (Figure~\ref{fig:general_fig1}). However, most of these models have been developed and evaluated on private datasets, which makes it difficult to assess their generalizability. ETAB (\cite{m2022etab}) provided an early benchmark with four public datasets, but its tasks overlap across datasets and views, and it focused on supervised adaptation rather than zero-shot or cross-modal evaluation. As a result, no standardized benchmark exists for modern echocardiography foundation models.

Beyond the lack of standardized benchmarks, echocardiography presents unique challenges for foundation model development. Unlike static radiographs, echocardiography produces temporal sequences of 2D images. Ultrasound images are inherently noisy and temporally complex, with high frame-to-frame similarity that makes effective representation learning more difficult (\cite{kang2024evaluation,song2024survey}). As shown in Figure~\ref{fig:general_fig1}, echocardiography videos exhibit much higher frame-level cosine similarity than natural videos, reflecting the low signal-to-noise ratio and limited visual variation of the modality. These characteristics have been associated with reduced robustness and limited generalization when models are trained directly on noisy ultrasound data (\cite{javed2024robustness}). They also make it hard to identify which modeling choices truly help, highlighting the need for systematic evaluation.

Furthermore, it remains unclear how these modality-specific models compare with general-purpose vision foundation models, which have much larger diversity in training data. This raises several fundamental questions: 1) How do echocardiography foundation models perform relative to each other under a fixed evaluation protocol? 2) Are their learned representation spaces fundamentally different from those of general-purpose models, and how do these differences affect downstream tasks? 3) To what extent can they enable zero-shot transfer, and do they exhibit systematic biases across datasets or clinical tasks? Addressing these open questions is essential for establishing reliable foundations for echocardiography AI, with direct implications for both methodological progress and the safe and reliable translation of these technologies into clinical practice. 

This work introduces \textbf{CardioBench} (see \autoref{fig:heart}), a comprehensive benchmark for echocardiography foundation models. By unifying eight publicly available datasets into a standardized evaluation suite spanning four regression and five classification tasks, CardioBench establishes the common ground for fair, reproducible, and clinically meaningful comparison. Unlike prior efforts that focused on individual datasets or tasks, CardioBench enables systematic evaluation across functional and structural endpoints, providing a robust basis for tracking progress in this emerging field. It compares leading cardiac-specific models with general-purpose vision and biomedical encoders under consistent zero-shot, probing, and alignment protocols, offering controlled analysis of how architectural design, temporal modeling, and supervision strategies shape transferability. To maximize accessibility and reproducibility, CardioBench provides standardized dataset preprocessing and data splits together with unified evaluation scripts, ensuring that results are directly comparable and easily extendable by the community.

Beyond results, CardioBench provides actionable insights into what drives performance in echocardiography foundation models: the role of temporal modeling, the importance of text encoders, the robustness of retrieval-based methods, and the surprising strengths and weaknesses of generalist backbones. We expect CardioBench to:  
(1) stimulate the development of new models tailored to the unique challenges of echocardiography,  
(2) establish a systematic way of measuring model quality for scientific progress, and  
(3) guide future pretraining strategies by revealing which architectural and supervision choices yield meaningful representations.

\section{Related work}
Recent works have advanced benchmarking and foundation models in medical imaging across multiple domains. \cite{bassi2024touchstone} builds a large-scale segmentation benchmark across nine abdominal organs to test models under distribution shift. Beyond performance, \cite{jin2024fairmedfm} emphasized fairness by assessing foundation models across multiple modalities and sensitive attributes. At the same time, \cite{huix2024natural} highlights the difficulty of transferring general-purpose foundation models to specialized modalities.

In echocardiography, \cite{m2022etab} provided an important early benchmark by assembling four public datasets into 31 tasks, establishing the first standardized protocol for model comparison. Many of these tasks, however, overlap across datasets and views, offering breadth but limited diversity in clinically distinct evaluation endpoints. Furthermore, ETAB primarily focused on supervised task adaptation, leaving a critical gap in evaluating the zero-shot, probing, and cross-modal alignment capabilities of modern multimodal foundation models. Since then, several echocardiography foundation models have been released, many of which are evaluated only on private datasets, which limits reproducibility and fair comparison across methods (\cite{song2024survey}). Together, these works highlight the absence of a standardized benchmark specifically tailored for echocardiography foundation models, underscoring the need for a public protocol that enables fair evaluation under noise and domain shifts in cardiac ultrasound.

\section{Methodology}

Cardiobench is a unified benchmark designed to evaluate echocardiography foundation models across multiple clinical tasks, datasets, and evaluation regimes under minimal adaptation. We assess both out-of-the-box performance and lightweight adaptation. We define zero-shot evaluation as inference without any training on the target dataset, using only model components released by the original authors. This setting reflects the intended use of foundation models, which are not retrained for every downstream task or deployment scenario. Together, these design choices provide a consistent framework for comparing models with diverse architectures and training strategies.

Within this framework, we design tasks that reflect the clinical use of echocardiography. Echocardiography provides a comprehensive view of the heart, capturing its motion, structure, and pathological states across time. To rigorously benchmark foundation models in this domain, we design tasks that capture functional, structural, and diagnostic aspects of clinical practice, as illustrated in Figure~\ref{fig:heart}. Functional tasks reflect the heart’s movement over time, with Left Ventricular Ejection Fraction (LV EF) regression serving as a standard measure of global cardiac performance that requires models to capture temporal dynamics across the cardiac cycle. Structural tasks emphasize the heart's anatomical properties, focusing on diastolic measurements (IVSd, LVIDd, LVPWd) to assess the spatial localization of cardiac walls. At the same time, diagnostic tasks focus on disease classification, including aortic stenosis (AS), pulmonary arterial hypertension (PAH), atrial septal defect (ASD), ST-elevation myocardial infarction (STEMI), and regional wall motion abnormality (RWMA) from 3 different views, thereby testing adaptability to diverse clinical targets. 

Beyond core tasks, the CardioBench also accounts for echocardiography’s broader context, including its multi-view nature and potential demographic biases. Echocardiography is inherently multi-view, with different pathologies and anatomical structures visible only from specific perspectives. View classification is therefore essential, as accurate recognition enables physicians to interpret the correct structures and ensures that automated models apply the appropriate downstream diagnostic tasks. In addition, we analyze demographic and acquisition-related factors, providing insight into subgroup robustness. 

For evaluation, we consider three categories of foundation models: those designed specifically for echocardiography, those trained on broader biomedical data, and large-scale general-purpose models. These span a wide range of architectural choices, from models without text supervision to those with temporal transformers over frame sequences or purely image-level extractors. Taken together, these variations in scale, architecture, and pretraining strategy allow us to assess how different design choices transfer to echocardiography interpretation (see Supplementary Material).

\textbf{Echocardiography–specific FM.} We evaluate the four Echocardiography foundation models with publicly released weights available at the time of writing. The earliest, EchoCLIP (\cite{christensen2024vision}), introduced a vision–language approach to cardiac ultrasound. EchoPrime (\cite{vukadinovic2024echoprime}) built on this idea with a stronger video encoder and a larger dataset, while also incorporating a separate view classifier and relying on report retrieval at inference time. In parallel, PanEcho (\cite{holste2025panecho}) explored an alternative direction by discarding text supervision and instead combining frame features with temporal aggregation in a multitask setup, while EchoFM (\cite{kim2024echofm}) explored a generative pretraining strategy centered on reconstructing cardiac motion. As it lacks a text encoder and the weights for its task-specific linear heads were not released, it cannot perform zero-shot evaluation. 

\textbf{Biomedical and general-purpose FM.} To assess transfer from broader domains, we also include BioMedCLIP (\cite{zhang2023BioMedCLIP}), a vision–language model pretrained on millions of biomedical image–text pairs spanning radiology, microscopy, pathology, and ultrasound. For comparison, we evaluate two large-scale general-purpose models trained at internet scale: DINOv3 (\cite{simeoni2025dinov3}), a self-supervised vision transformer, and SigLIP2 (\cite{tschannen2025siglip}), a vision–language model designed to produce stronger dense representations. Together, these models enable testing of how far biomedical and generic pretraining can transfer to echocardiography tasks, and whether domain-specific pretraining is required to achieve strong performance.

%%% ZERO SHOT TABLE 
\begin{table*}[t]
\centering
\setlength{\tabcolsep}{6pt}
\renewcommand{\arraystretch}{1.2}
\resizebox{\linewidth}{!}{%
\begin{tabular}{l|c|c|c|c|c|c|c|c|c|c|c|c|c}
\hline
\multirow{2}{*}{\textbf{Model}} &
\textbf{Dynamic} & \textbf{CAMUS} & \textbf{Pediatric} &
\multicolumn{3}{c|}{\textbf{LVH}} &
\multicolumn{2}{c|}{\textbf{CardiacNet}} &
\textbf{HMC-QU} & \textbf{TMED-2} &
\multicolumn{3}{c}{\textbf{segRWMA}} \\
\cline{2-14} 
& \multicolumn{3}{c|}{LV EF (\%) $\downarrow$} &
LVIDd (cm) $\downarrow$ & IVSd (cm) $\downarrow$ & LVPWd (cm) $\downarrow$ &
ASD $\uparrow$ & PAH $\uparrow$ &
STEMI $\uparrow$ & AS $\uparrow$ &
A2C $\uparrow$ & A3C $\uparrow$ & A4C $\uparrow$ \\
\Xhline{1pt}
\textbf{EchoCLIP} \cite{christensen2024vision} & 
9.99 & \underline{9.83} & 13.80 & 0.79 & 0.57 & 0.41 & \textbf{47.88} & \textbf{46.96} & \textbf{52.51} & \underline{44.13} & 35.68 & 36.27 & 14.29 \\
\textbf{EchoPrime} \cite{vukadinovic2024echoprime} \bluelozenge & \underline{7.78} & 14.00 & \textbf{5.44} & - & - & - & - & - & - & \underline{44.13} & - & - & - \\
\textbf{PanEcho} \cite{holste2025panecho} \bluelozenge & \textbf{5.79} & 11.63 & \underline{9.10} & \textbf{0.36} & \textbf{0.21} & \textbf{0.18} & - & - & - & \textbf{58.90} & 30.50 & 24.30 & 20.52 \\
\textbf{BioMedCLIP} \cite{zhang2023BioMedCLIP} & 13.83 & 18.87 & 18.30 & 0.97 & \underline{0.28} & 0.26 & \underline{40.24} & 25.75 & 33.33 & \underline{44.13} & 37.66 & 32.10 & 6.67 \\
\textbf{DINOv3} \cite{simeoni2025dinov3} & 14.67 & 9.88 & 18.24 & \underline{0.69} & \underline{0.28} & \underline{0.22} & 36.49 & \underline{41.44} & \underline{34.21} & \underline{44.13} & \textbf{47.83} & \underline{48.00} & \textbf{48.15} \\
\textbf{SigLIP2} \cite{tschannen2025siglip} & 14.66 & \textbf{9.28} & 18.22 & \underline{0.69} & \underline{0.28} & \underline{0.22} & 36.49 & 24.11 & 32.43 & 17.38 & \underline{47.25} & \textbf{72.02} & \underline{47.17} \\
\hline
\end{tabular}
}
\caption{Zero-shot results across 4 regression tasks and 5 classification tasks on 8 publicly available datasets. \textbf{Regression tasks} (reported in MAE, lower is better $\downarrow$): LV EF on Dynamic~\cite{echonetdynamic2019}, CAMUS~\cite{leclerc2019deep}, and Pediatric~\cite{echonetpediatric2025}; LVIDd, IVSd, and LVPWd on LVH~\cite{echonetlvh2020}. \textbf{Classification tasks} (reported in F1-macro, higher is better $\uparrow$): ASD and PAH on CardiacNet~\cite{yang2024cardiacnet}; STEMI on HMC-QU~\cite{degerli2021early}; AS on TMED-2~\cite{huang2022tmed}; and A2C, A3C, and A4C on segRWMA~\cite{liu2023enhance}. Models with video-based training are marked with \bluelozenge. The best results are shown in \textbf{bold}, and the second best are \underline{underlined}.}
\label{tab:Zeroshot}
\end{table*}
% %%% ZERO SHOT TABLE 

%%% LINEAR PROBING TABLE
%%% LINEAR PROBING TABLE
\begin{table*}[t]
\centering
\setlength{\tabcolsep}{4pt} % tighten spacing for paired columns
\renewcommand{\arraystretch}{1.2}
\resizebox{\linewidth}{!}{%
\begin{tabular}{l|c c|c c|c c|c c|c c|c c|c c|c c|c c}
\hline
\multirow{2}{*}{\textbf{Model}} &
\multicolumn{6}{c|}{\textbf{LVH}} &
\multicolumn{4}{c|}{\textbf{CardiacNet}} &
\multicolumn{2}{c|}{\textbf{HMC-QU}} &
\multicolumn{6}{c}{\textbf{segRWMA}} \\
\cline{2-19}
& LVIDd (cm) $\downarrow$ & $\Delta$ & IVSd (cm) $\downarrow$ & $\Delta$ & LVPWd (cm) $\downarrow$ & $\Delta$ &
ASD $\uparrow$ & $\Delta$ & PAH $\uparrow$ & $\Delta$ &
STEMI $\uparrow$ & $\Delta$ &
A2C $\uparrow$ & $\Delta$ & A3C $\uparrow$ & $\Delta$ & A4C $\uparrow$ & $\Delta$ \\
\Xhline{1pt}
\textbf{EchoCLIP} \cite{christensen2024vision}  
& 0.47 & \up{0.32} & 0.28 & \up{0.29} & 0.22 & \up{0.19} 
& 38.49 & \down{9.39} & 41.44 & \down{5.52} & 73.99 & \up{21.48} 
& \underline{47.83} & \up{12.15} & \underline{48.00} & \up{11.73} & \underline{48.15} & \up{38.86} \\
\textbf{EchoPrime} \cite{vukadinovic2024echoprime} \bluelozenge
& \underline{0.41} & – & \underline{0.25} & – & \underline{0.19} & – 
& 52.66 & – & \textbf{63.36} & – & \textbf{80.00} & – 
& 8.33 & – & \textbf{68.48} & – & \underline{48.15} & – \\
\textbf{PanEcho} \cite{holste2025panecho} \bluelozenge 
& \textbf{0.35} & \up{0.01} & \textbf{0.18} & \up{0.03} & \textbf{0.15} & \up{0.03} 
& \underline{58.53} & – & \underline{61.51} & – & 69.70 & – 
& \textbf{72.73} & \up{42.23} & 47.47 & \up{23.17} & \textbf{64.78} & \up{44.26} \\
\textbf{EchoFM} \cite{kim2024echofm}   
& 0.57 & – & 0.32 & – & 0.24 & – 
& 50.48 & – & 41.44 & – & 71.82 & – 
& \underline{47.83} & – & \underline{48.00} & – & \underline{48.15} & – \\
\textbf{BioMedCLIP} \cite{zhang2023BioMedCLIP} 
& 0.52 & \up{0.45} & 0.30 & \down{0.02} & 0.23 & \up{0.03} 
& \underline{58.53} & \up{1.20} & 41.44 & \up{15.69} & 55.44 & \up{22.11} 
& \underline{47.83} & \up{10.17} & \underline{48.00} & \up{15.90} & \underline{48.15} & \up{41.48} \\
\textbf{DINOv3} \cite{simeoni2025dinov3}     
& 0.47 & \up{0.22} & 0.28 & \up{0.00} & 0.21 & \up{0.01} 
& 56.76 & \up{22.36} & 58.85 & \up{17.41} & \underline{75.00} & \up{40.79} 
& \underline{47.83} & \up{0.00} & \underline{48.00} & \up{0.00} & \underline{48.15} & \up{0.00} \\
\textbf{SigLIP2} \cite{tschannen2025siglip}    
& 0.51 & \up{0.18} & 0.30 & \down{0.02} & 0.23 & \down{0.01} 
& \textbf{68.49} & \up{32.00} & 47.96 & \up{23.85} & \underline{75.00} & \up{42.57} 
& \underline{47.83} & \up{0.48} & \underline{48.00} & \down{24.02} & \underline{48.15} & \up{0.98} \\
\hline
\end{tabular}}
\caption{Linear probing results across 3 regression tasks and 4 classification tasks on 4 publicly available datasets. \textbf{Regression tasks} (reported in MAE, lower is better $\downarrow$): LVIDd, IVSd, and LVPWd on LVH. \textbf{Classification tasks} (reported in F1-macro, higher is better $\uparrow$): ASD and PAH on CardiacNet; STEMI on HMC-QU; and A2C, A3C, and A4C on segRWMA. Reported $\Delta$ values indicate absolute change relative to zero-shot. Models with video-based training are marked with \bluelozenge. The best results are shown in \textbf{bold}, and the second best are \underline{underlined}.}
\label{tab:LinearProbingDelta}
\end{table*}
% \section{Experiments}
% \subsection{Experimental design principles}
We design experiments to examine two complementary aspects of foundation models: \textbf{(i) the capacity to perform clinically relevant tasks without task-specific training}, and \textbf{(ii) the quality of their learned representations for downstream adaptation}. Therefore, we focus on zero-shot evaluation and probing, while excluding fine-tuning and few-shot training, as both are prone to overfitting and require substantial labeled data for stable performance (\cite{silva2024closer}). Further details on zero-shot evaluation, prompt design, and probing implementations are provided in Supplementary Material.

Foundation models are evaluated on both predictive accuracy and the structure of their learned representations. We therefore report metrics across four dimensions: task performance, clustering consistency, cross-modal alignment, and demographic robustness. For task performance, we use Mean Absolute Error (MAE) as the primary regression metric and macro-averaged F1 for classification and view classification. Clustering consistency is assessed using the Adjusted Rand Index (ARI), which measures how well embedding clusters recover ground-truth echocardiography views. Cross-modal alignment is evaluated by testing whether visual embeddings align with text prompts. Finally, demographic robustness is examined through subgroup analyses of EF errors stratified by sex, age, BMI, and image quality.

\section{Results}

%%%%% VIEW CLASSIFICATION TABLE
\begin{table*}[t]
\centering
\setlength{\tabcolsep}{6pt}
\renewcommand{\arraystretch}{1.2}
\resizebox{\linewidth}{!}{%
\begin{tabular}{l|c|c|c|c|c|c|c|c}
\hline
\textbf{Model} & \textbf{LVH} $\uparrow$ & \textbf{CardiacNet} $\uparrow$ &
\textbf{CAMUS} $\uparrow$ \orangelozenge & \textbf{Dynamic} $\uparrow$ \orangelozenge & \textbf{Pediatric} $\uparrow$ \orangelozenge &
\textbf{HMC-QU} $\uparrow$ \orangelozenge & \textbf{TMED-2} $\uparrow$ \orangelozenge & \textbf{segRWMA} $\uparrow$ \orangelozenge \\
\Xhline{1pt}
\textbf{EchoCLIP} \cite{christensen2024vision}   & 1.76 & 27.12 & \underline{33.11} & 8.55 & 20.95 & 34.33 & 14.25 & \underline{16.86} \\
\textbf{EchoPrime} \cite{vukadinovic2024echoprime} & \textbf{98.66} & \underline{34.59} & 16.39 & \textbf{98.49} & \textbf{79.53} & \textbf{88.19} & \textbf{62.86} & 15.79 \\
\textbf{BioMedCLIP} \cite{zhang2023BioMedCLIP}   & 0.57 & \textbf{76.11} & 17.02 & 26.37 & 18.41 & \underline{47.67} & \underline{21.98} & \textbf{18.41} \\
\textbf{DINOv3} \cite{simeoni2025dinov3}         & 0.00 & 0.00 & 0.00 & 0.31 & 35.82 & 0.00 & 4.89 & 0.00 \\
\textbf{SigLIP2} \cite{tschannen2025siglip}      & \underline{29.05} & 8.75 & \textbf{57.01} & \underline{87.29} & \underline{45.32} & 41.37 & 16.17 & 2.43 \\
\hline
\end{tabular}
}
\caption{View classification results across 8 publicly available datasets, reported in \textbf{F1-macro} score. Multi-view datasets are marked with \orangelozenge. The best results are shown in \textbf{bold}, and the second best are \underline{underlined}.}
\label{tab:ViewClassification}
\end{table*}
We summarize the performance of models in a zero-shot setting in Table~\ref{tab:Zeroshot}. PanEcho is the most consistent performer, achieving the best and second-best results for ejection fraction (EF) estimation on EchoNet-Dynamic and EchoNet-Pediatric, and outperforming all competitors on the structural regression tasks from EchoNet-LVH. Its strength extends to classification, where it achieves the highest score of 58.90\% on TMED-2 aortic stenosis (AS) detection. EchoPrime shows strong results in both regression and classification tasks, which is particularly interesting given its retrieval-based inference framework and the potential influence of similarities between test cases and its private database.

A notable observation is the performance of general-purpose foundation models such as SigLIP2 and DINOv3, which deliver strong results despite lacking cardiac-specific pretraining. SigLIP2, in particular, surpasses several specialized echocardiography models on CAMUS EF estimation and achieves competitive performance on segRWMA regional wall abnormality detection. At the same time, both SigLIP2 and DINOv3 perform nearly on par with PanEcho on EchoNet-LVH regression LVPWd. In classification, they achieve the highest scores in RWMA detection across all three views, even outperforming EchoCLIP, despite EchoCLIP being explicitly trained on cardiac ultrasound. This underperformance is most pronounced on the A4C view, where EchoCLIP lags by more than 34\%. Nevertheless, EchoCLIP remains strong on several tasks, achieving F1 scores of 47.88\% on ASD and 46.96\% on PAH, surpassing the best general-purpose models by margins of 7.61\% and 5.52\%, respectively. On STEMI detection, it reaches 52.51\%, representing an improvement of 18.3\% over competitors. 

The linear probing performance is summarized in Table~\ref{tab:LinearProbingDelta}. On regression tasks, PanEcho maintains a clear advantage, achieving the lowest errors across all EchoNet-LVH measurements (MAE of 0.35 on LVIDd, 0.15 on IVSd, and 0.30 on LVPWd), with only marginal improvements from linear probing ($\Delta \leq 0.03$). By contrast, general-purpose encoders such as DINOv3 and SigLIP2 show larger reductions in error ($0.20$–$0.23$ MAE), narrowing the gap to PanEcho, though they remain behind. These results illustrate that EchoNet-LVH structural regression benefits less from probing. For classification, linear probing yields more pronounced changes. SigLIP2 improves by 32\% on ASD to reach 68.49\% F1, outperforming all specialized models by nearly 10\%. On PAH and STEMI, however, EchoPrime delivers the strongest performance, achieving 63.36\% and 80.00\%, while SigLIP2 remains competitive at 47.96\% and 72.57\%, respectively. These results show that general-purpose encoders can not only close the gap but, in some cases, even surpass specialized models.

In RWMA detection, PanEcho achieves the highest gains, with improvements of 42.23\% on A2C and 44.26\% on A4C, reaching 72.73\% and 64.78\%, respectively. EchoPrime excels on A3C, where it reaches 68.48\%, while EchoCLIP remains flat at 48.00\% across all views, converging with DINOv3 and SigLIP2 despite its cardiac-specific training. Overall, linear probing highlights complementary strengths with PanEcho remaining unrivaled on regression and two RWMA views, EchoPrime achieving the best results on PAH and STEMI, and SigLIP2 surpassing all competitors on ASD.

View classification results in Table~\ref{tab:ViewClassification} show that EchoPrime achieves the highest F1 scores on the majority of datasets, benefiting from its supervised, pretrained view classifier rather than relying solely on text–prompt alignment. It leads on five out of eight datasets, highlighting the strength of its dedicated view recognition module. Interestingly, the remaining datasets are topped by models without cardiac-specific pretraining: BioMedCLIP achieves the best results on CardiacNet (76.11\%) and TMED-2 (62.86\%), while SigLIP2 outperforms all others on CAMUS (57.01\%). For general-purpose models, however, view classification can break down when learned representations do not sufficiently separate visually similar echocardiographic views, leading to near-zero macro-F1 on some datasets. EchoCLIP, despite being trained specifically on echocardiography, fails to dominate on any dataset and often lags behind BioMedCLIP or general-purpose models. These findings suggest that while supervised view classifiers provide a clear advantage, large-scale pretraining on diverse medical or natural images can transfer surprisingly well to echocardiography view classification.

\begin{figure}[ht] % Single column figure
    \centering
    
    % Row 1: Sex
    \begin{subfigure}[b]{0.49\linewidth}
        \centering
        \includegraphics[width=\linewidth]{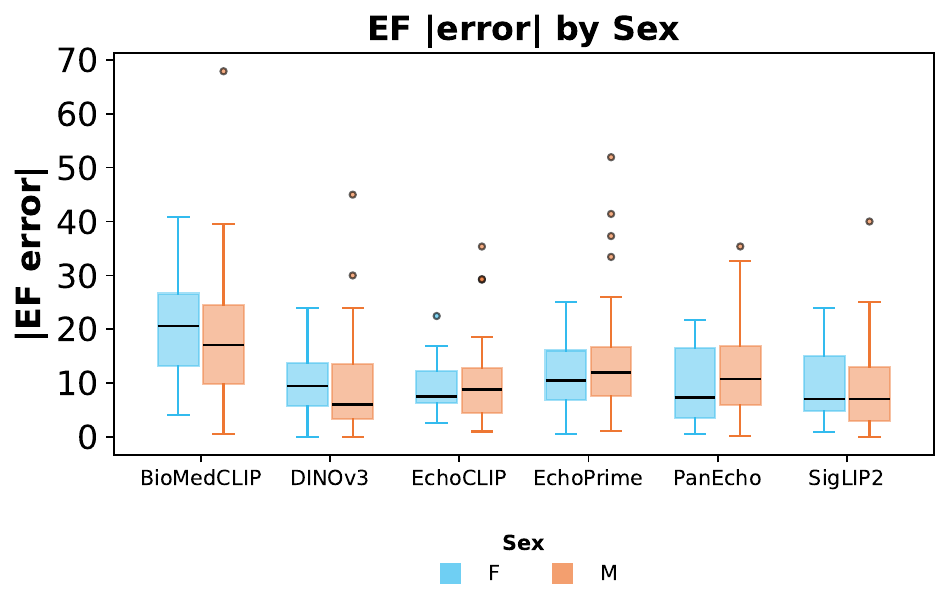}
        \caption{CAMUS: Sex}
        \label{fig:camus_sex}
    \end{subfigure}
    \hfill
    \begin{subfigure}[b]{0.48\linewidth}
        \centering
        \includegraphics[width=\linewidth]{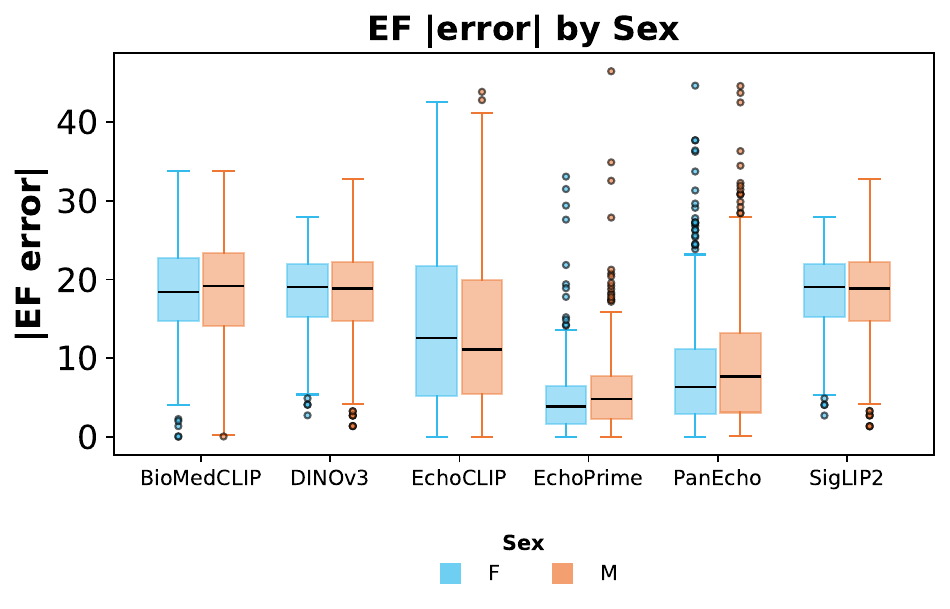}
        \caption{Pediatric: Sex}
        \label{fig:peds_sex}
    \end{subfigure}

    \vspace{0.5em}

    % Row 2: Age
    \begin{subfigure}[b]{0.49\linewidth}
        \centering
        \includegraphics[width=\linewidth]{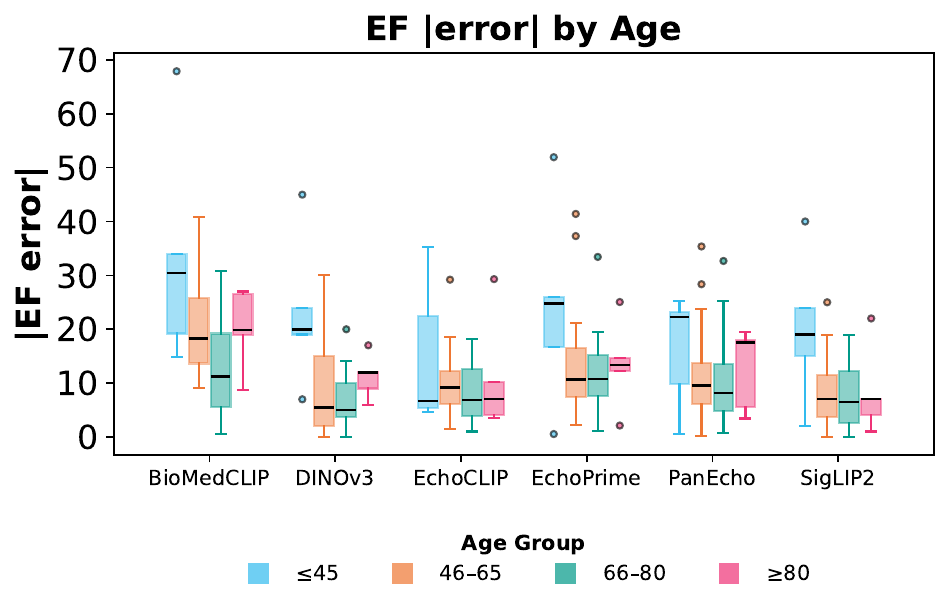}
        \caption{CAMUS: Age}
        \label{fig:camus_age}
    \end{subfigure}
    \hfill
    \begin{subfigure}[b]{0.48\linewidth}
        \centering
        \includegraphics[width=\linewidth]{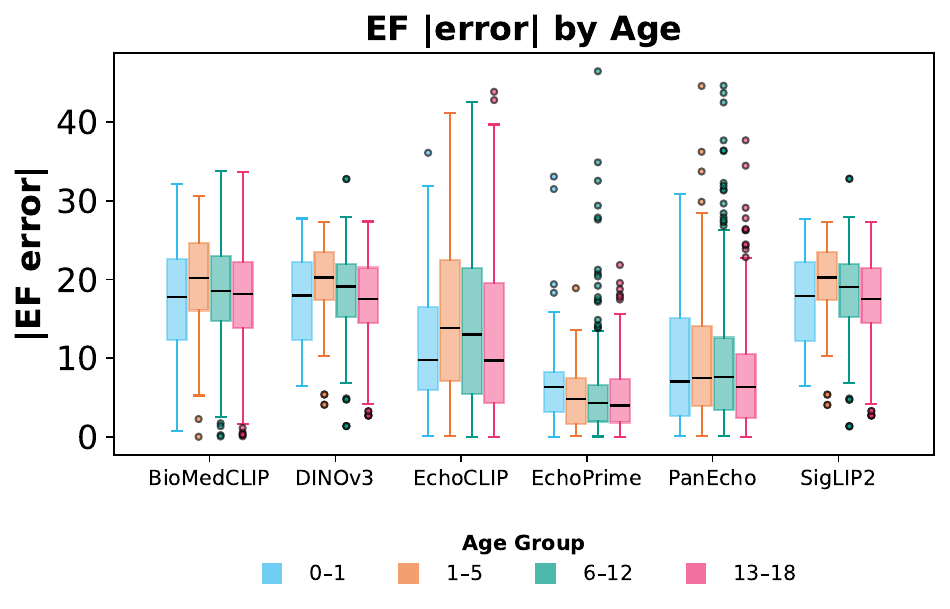}
        \caption{Pediatric: Age}
        \label{fig:peds_age}
    \end{subfigure}

    \vspace{0.5em}

    % Row 3: Image Quality / BMI
    \begin{subfigure}[b]{0.49\linewidth}
        \centering
        \includegraphics[width=\linewidth]{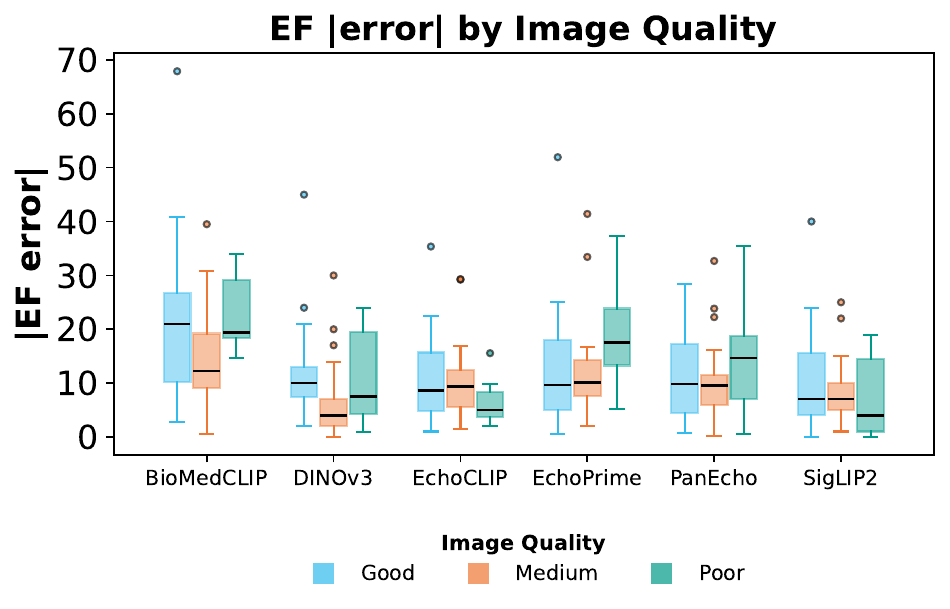}
        \caption{CAMUS: Quality}
        \label{fig:camus_quality}
    \end{subfigure}
    \hfill
    \begin{subfigure}[b]{0.48\linewidth}
        \centering
        \includegraphics[width=\linewidth]{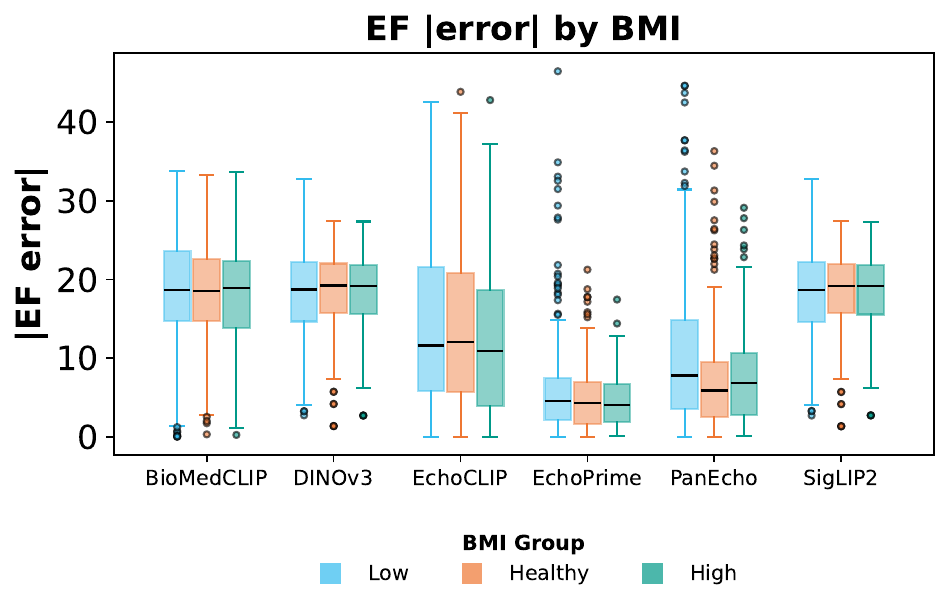}
        \caption{Pediatric: BMI}
        \label{fig:peds_bmi}
    \end{subfigure}

    \caption{Absolute EF error distributions by subgroup: Sex (top), Age (middle), and Image Quality/BMI (bottom).}
    \label{fig:demographics}
\end{figure}

Subgroup analyses reveal distinct biases in EF estimation on CAMUS that are less pronounced in EchoNet-Pediatric (Figure~\ref {fig:demographics}), despite overall performance trends being consistent across models. On CAMUS (Figure~\ref{fig:demographics}a,c,e), subgroup differences are evident: younger patients ($\leq$45) and scans labeled as “Good” quality show larger errors and wider spreads, likely reflecting distribution biases since most samples fall into the “Medium” quality category, where models perform best. A modest sex gap is also visible, with females showing slightly higher errors, particularly for EchoPrime and PanEcho. In the larger EchoNet-Pediatric cohort (Figure~\ref{fig:demographics}b,d,f), these disparities are less pronounced. Sex- and age-related differences largely disappear, while BMI exhibits the expected trend: healthy ranges yield lower errors, whereas both low and high extremes increase variability, consistent with the physics of ultrasound imaging, where excessive or insufficient tissue layers can degrade acoustic penetration and image quality. Across both datasets, SigLIP2 and DINOv3 maintain the most stable performance across demographic and acquisition subgroups, showing narrow error distributions and minimal subgroup-related shifts. BioMedCLIP, while consistently higher in absolute error, also shows relatively uniform behavior across subgroups. By contrast, PanEcho and EchoPrime demonstrate more outliers and wider error distributions across several subgroups, particularly in females and younger patients on CAMUS and in BMI extremes on EchoNet-Pediatric.

% We evaluate model robustness across patient demographics and image quality, observing that while performance trends are generally consistent, specific subgroups like younger patients or BMI extremes exhibit higher error variability due to dataset biases and ultrasound physics. General-purpose encoders like SigLIP2 and DINOv3 demonstrate the most stable performance with minimal subgroup-related shifts compared to specialized models (see Appendix~\ref{app:demographics}).

\section{Discussion}
CardioBench reveals that no single foundation model dominates across all tasks, datasets, and evaluation regimes. Instead, performance depends strongly on the interaction between model design choices, dataset characteristics, and evaluation setup.
\begin{figure}[t] 
    \centering
    \includegraphics[width=\columnwidth]{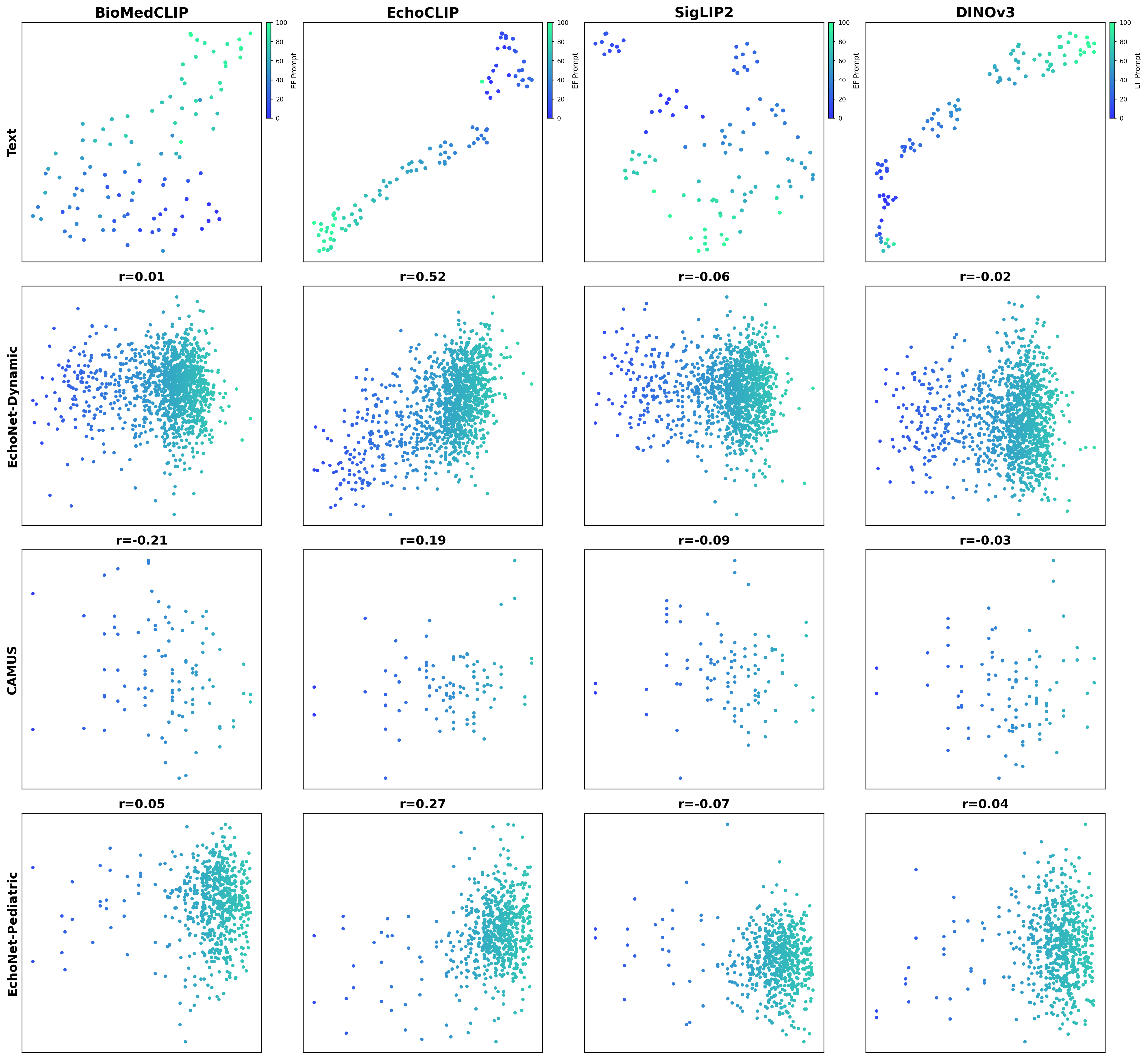} 
    \caption{Top row: EF text prompt embeddings projected into 2D. Rows 2--4: alignment of visual embeddings with the EF text axis for each dataset.}
    \label{fig:ef_text_align}
\end{figure}
% \begin{figure}[t] 
%     \centering
%     \includegraphics[width=\columnwidth]{figures/discussion/ef_visuals.png}
%     \caption{UMAP of visual representations on EchoNet-Dynamic, CAMUS, and EchoNet-Pediatric Datasets across the four foundation models. }
%     \label{fig:ef_visuals}
% \end{figure}
% \begin{figure*}[t]
%     \centering
%     % First Subfigure (Left)
%     \begin{subfigure}{0.3\textwidth}
%         \centering
%         \includegraphics[width=\linewidth]{figures/discussion/text_align_ef.png}
%         \caption{EF text prompt embeddings projected into 2D. Rows 2--4: alignment of visual embeddings with the EF text axis for each dataset.}
%         \label{fig:ef_text_align}
%     \end{subfigure}
%     \hfill 
%     \begin{subfigure}{0.68\textwidth}
%         \centering
%         \includegraphics[width=\linewidth]{figures/discussion/ef_visuals.png}
%         \caption{UMAP of visual representations on EchoNet-Dynamic, CAMUS, and EchoNet-Pediatric Datasets across the four foundation models.}
%         \label{fig:ef_visuals}
%     \end{subfigure}
    
%     \caption{Comparison of text alignment and visual representations across cardiac datasets.}
%     \label{fig:combined_ef_analysis}
% \end{figure*}
\textbf{Modeling EF regression.} PanEcho and EchoPrime stand apart from the contrastive approaches in CardioBench because their zero-shot predictions are not driven by text encoders. PanEcho leverages its multitask design to achieve the lowest errors on EchoNet-Dynamic and strong results on Pediatric, showing that supervised EF knowledge can transfer effectively across datasets. EchoPrime, in contrast, benefits from retrieval: rather than modeling EF as a smooth continuum, it assigns labels by matching test cases to similar exemplars in its joint space. This discrete matching helps on EchoNet-Pediatric, where it outperforms contrastive models, but the approach fails on CAMUS, where scanner heterogeneity may distort embeddings and make nearest-neighbor matches unreliable. Both models incorporate temporal dynamics, but differ in how strongly their predictions depend on them. A frame-shuffling stress test highlights this contrast: shuffling frames increases PanEcho MAE from 5.79 to 6.32, indicating sensitivity to temporal order, whereas EchoPrime remains relatively stable (7.79 to 6.12), suggesting that its retrieval mechanism can fall back on exemplar similarity even when sequence order is disrupted.

To examine contrastive approaches, we directly assess whether they encode EF as a cross-modal dimension. We construct a text axis from prompts spanning 0--100\% EF, normalize these embeddings, and extract the first principal component. Visual embeddings from test videos are then projected onto this axis, and their Pearson correlation with ground-truth EF quantifies alignment (Figure~\ref{fig:ef_text_align}). This analysis reveals significant differences between models. EchoCLIP, trained on cardiac ultrasound reports, is the only model to recover a physiologically meaningful EF axis ($r = 0.52$ on EchoNet-Dynamic, $r \approx 0.2$--$0.3$ on CAMUS and Pediatric), suggesting that domain-specific text encoders can enforce monotonic cross-modal structure. BioMedCLIP, despite pretraining on extensive biomedical corpora, shows almost no alignment ($r \approx 0$), indicating that general medical semantics are insufficient to ground EF as a continuous variable. General-purpose models such as SigLIP2 and DINOv3 also result in near-zero correlations, yet achieve their strongest results on CAMUS. At first glance, this might suggest robustness to acquisition shifts; however, a closer look indicates that these gains are not physiologically grounded. Specifically, we observe that SigLIP2 achieves lower MAE on images with poor quality compared to those of higher quality (Figure~\ref{fig:camus_quality}), which is counterintuitive from a clinical perspective. This pattern suggests that the apparent success of generalist models on CAMUS reflects sensitivity to dataset-specific artifacts rather than meaningful encoding of EF, explaining their poor generalization outside this narrow setting.

\begin{figure}[t]
    \centering
    % Change width to ~0.48 to allow space for the horizontal fill
    \begin{subfigure}{0.5\linewidth}
        \centering
        \includegraphics[width=\linewidth]{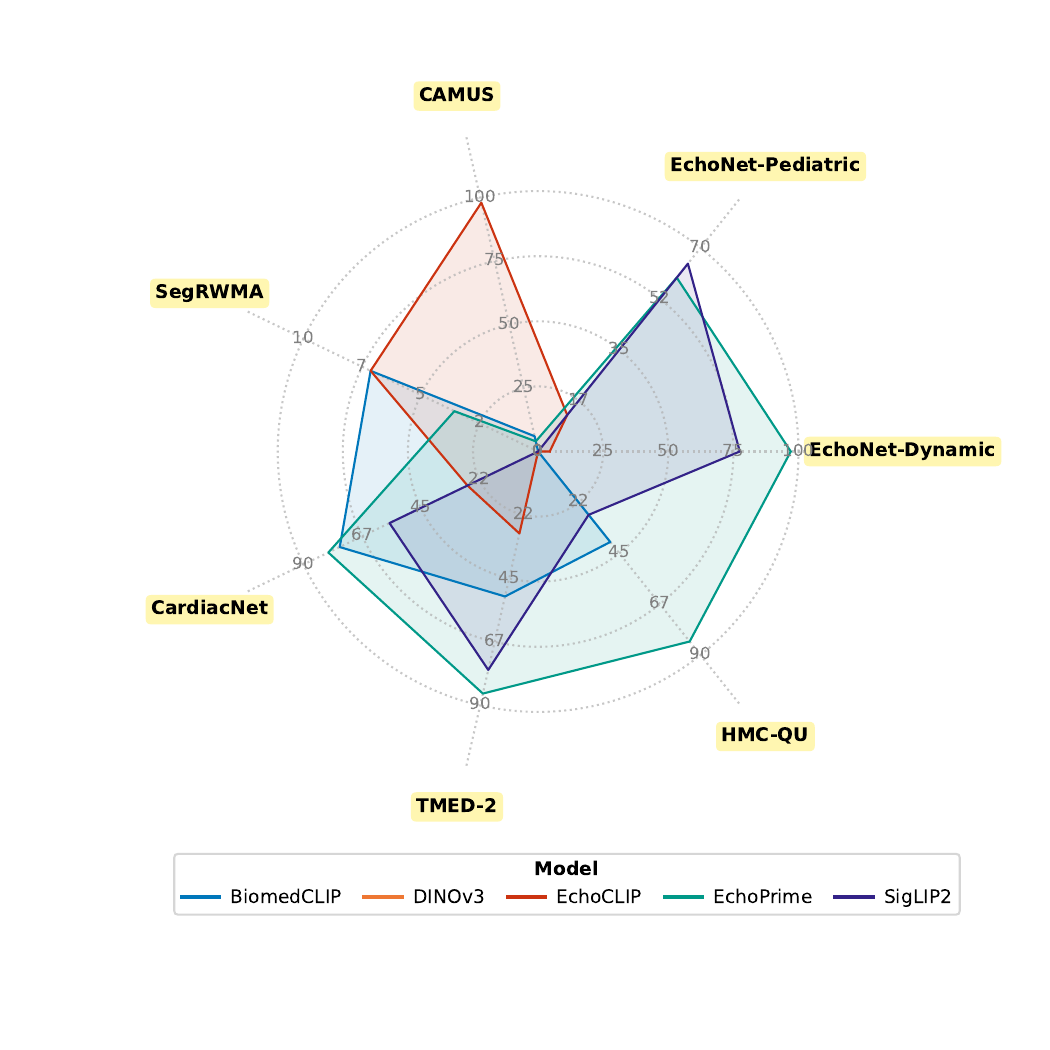}
        \caption{A4C View}
        \label{fig:radar_a4c}
    \end{subfigure}
    \hfill % This pushes the two images to the far left and right
    \begin{subfigure}{0.5\linewidth}
        \centering
        \includegraphics[width=\linewidth]{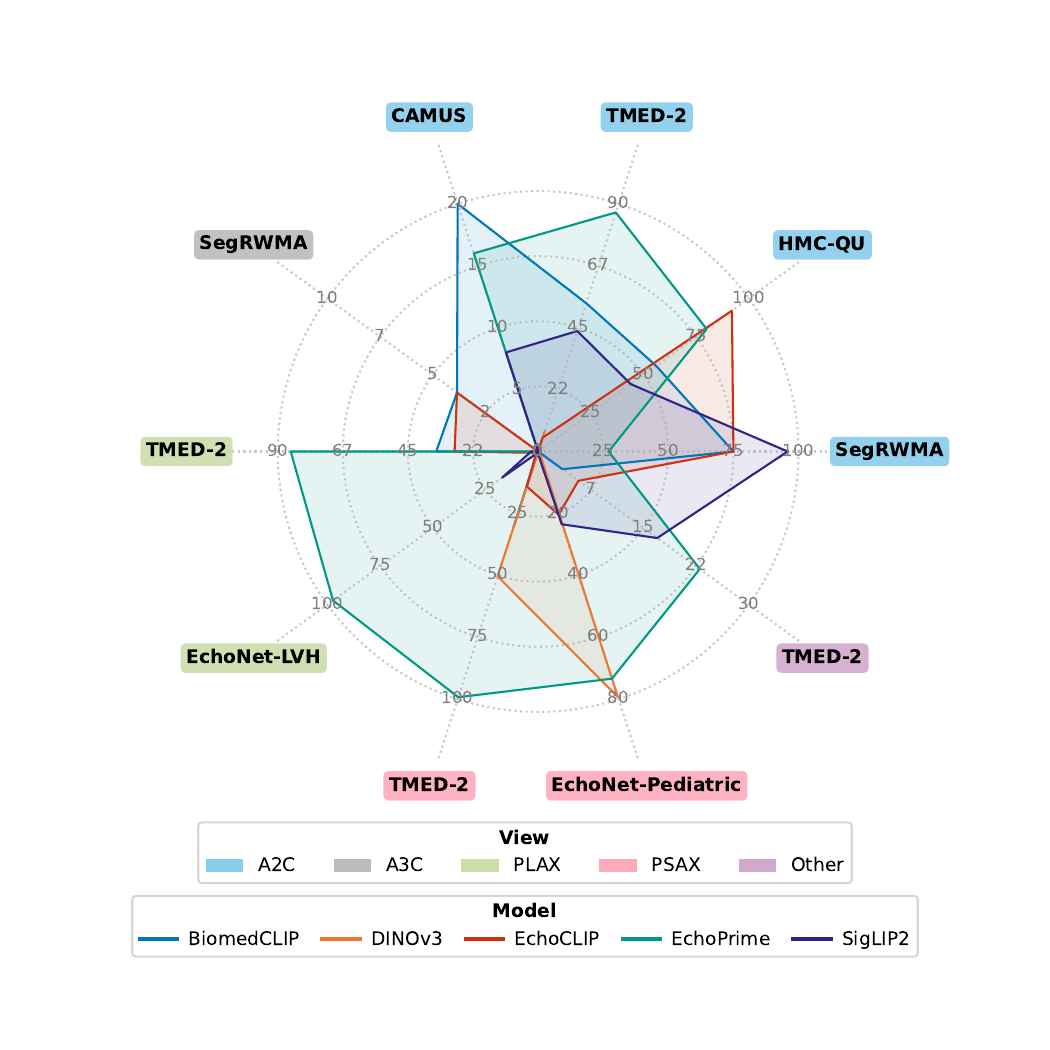}
        \caption{Other Views}
        \label{fig:radar_different}
    \end{subfigure}

    \caption{Comparison of view classification accuracy across different datasets using radar plots.}
    \label{fig:combined_radars}
\end{figure}

\begin{figure}[t]
    \centering
    \includegraphics[width=\linewidth]{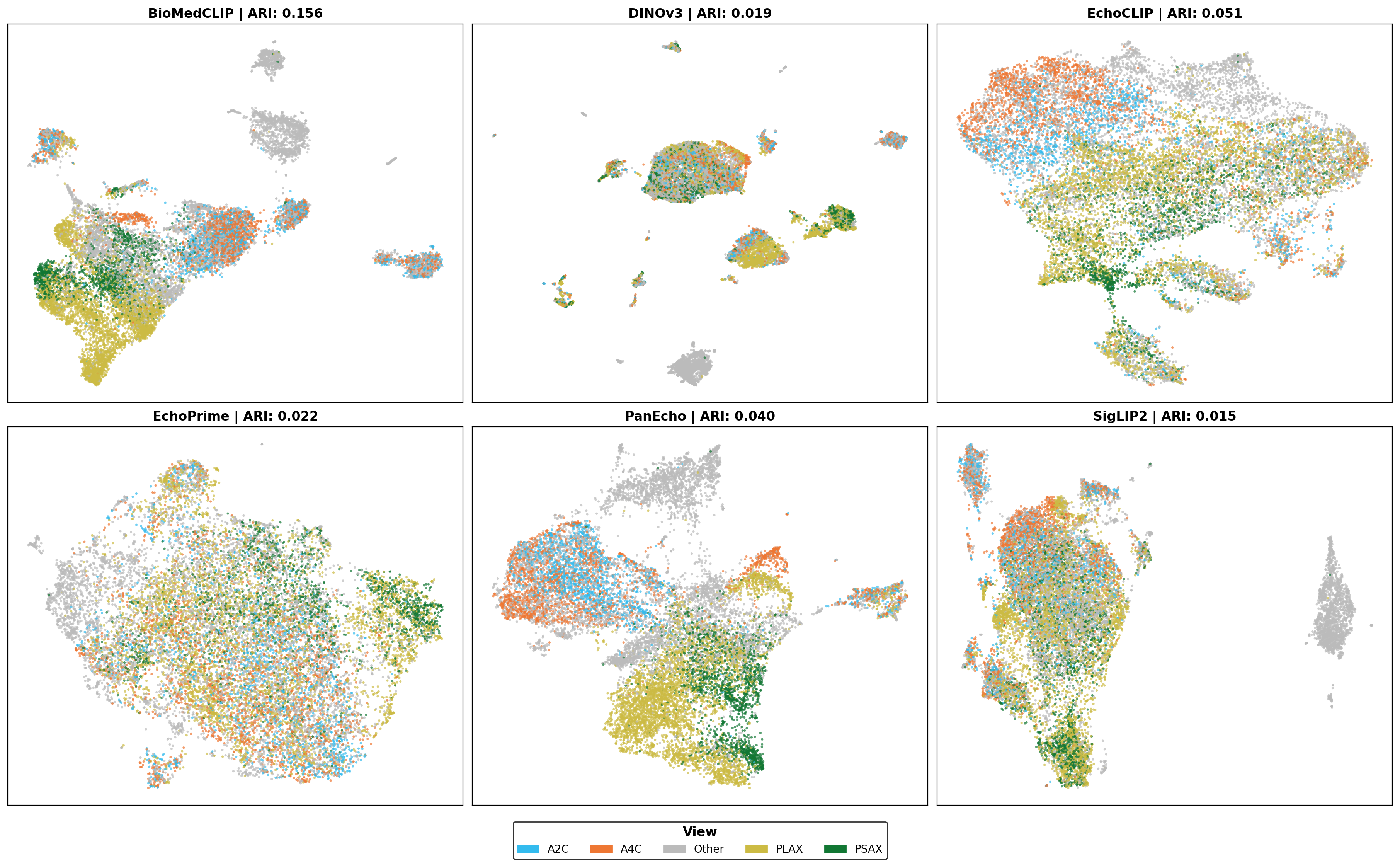}
    \caption{UMAP projection of TMED-2 embeddings highlighting view-based clustering.}
    \label{fig:umap_single}
\end{figure}
\begin{table}[t]
    \centering
    \scriptsize % Use scriptsize to fit many columns in one column width
    \setlength{\tabcolsep}{1.2pt} % Reduce padding between columns
    \renewcommand{\arraystretch}{1.2}
    
    \begin{tabular}{l|cc|cccccccccc}
        \toprule
        \textbf{Model} & \textbf{F1} & $\Delta$ & \textbf{A2C} & $\Delta$ & \textbf{A4C} & $\Delta$ & \textbf{PLAX} & $\Delta$ & \textbf{PSAX} & $\Delta$ & \textbf{Oth.} & $\Delta$ \\
        \midrule
        EchoCLIP   & 59.98 & \up{45.7} & 53.9 & \up{0.0} & \underline{48.1} & \up{19.1} & 62.7 & \up{27.5} & \underline{84.6} & \up{54.1} & 49.1 & \up{45.6} \\
        EchoPrime  & 50.87 & \down{12.0} & 37.7 & \down{49.2} & 39.8 & \down{46.0} & \underline{70.1} & \down{15.4} & 66.6 & \down{32.6} & 33.2 & \up{10.1} \\
        PanEcho    & \underline{63.46} & -- & \underline{57.6} & -- & 47.7 & -- & 66.0 & -- & 84.1 & -- & \textbf{70.2} & -- \\
        BioMedCLIP & \textbf{68.13} & \up{41.8} & \textbf{59.6} & \up{54.6} & \textbf{57.2} & \up{5.8} & \textbf{71.4} & \up{42.5} & \textbf{88.6} & \up{74.5} & \underline{66.3} & \up{60.6} \\
        DINOv3     & 50.92 & \up{46.0} & 36.7 & \up{29.3} & 43.3 & \up{42.8} & 61.0 & \up{61.0} & 75.9 & \up{25.5} & 40.0 & \up{39.8} \\
        SigLIP2    & 57.34 & \up{41.2} & 42.1 & \down{1.7} & 47.4 & \down{30.0} & 69.7 & \up{67.3} & 79.1 & \up{78.6} & 42.6 & \up{25.6} \\
        \bottomrule
    \end{tabular}
    
    \caption{KNN probing results on TMED-2, reporting F1-macro scores and per-view accuracies. $\Delta$ denotes the absolute change relative to zero-shot performance. The best results are \textbf{bold}, and the second-best are \underline{underlined}.}
    \label{tab:kmm_view}
\end{table}

\textbf{Clustering challenges in view classification.} A similar picture emerges in view classification, where architectural choices again dominate over text alignment. EchoPrime achieves the strongest results across multiple datasets by leveraging its supervised view head, demonstrating that explicitly modeling clinical structure can result in zero-shot advantages. By contrast, EchoCLIP struggles to generalize beyond A4C despite being trained on this view, because its contrastive objective emphasizes alignment with reports rather than enforcing consistent view identity. As a result, its embeddings entangle clinical content with anatomical cues, limiting transfer even on its main training view. Large-scale encoders such as BioMedCLIP and SigLIP2 occasionally outperform specialized models on datasets like EchoNet-Pediatric and CAMUS, but UMAP projections (Figure \ref{fig:umap_single}) of TMED-2 embeddings reveal that none of the models form globally distinct view clusters. Interestingly, BioMedCLIP, EchoCLIP, and PanEcho, which were not explicitly trained for view classification, tend to group PLAX and PSAX together while mixing A2C and A4C, as these views are indeed visually similar within short-axis and long-axis families. kNN probing (Table~\ref{tab:kmm_view}) recovers some discriminative power, ranking BioMedCLIP highest, followed by PanEcho and EchoCLIP, while SigLIP2 surpasses EchoPrime when its supervised view classifier is removed. This shows that EchoPrime’s advantage comes almost entirely from its explicit classifier head, while other models contain partial view information in their embeddings that kNN can recover locally, but which does not form globally distinct clusters or generalize consistently across datasets.

\begin{figure}[t]
    \centering
    \includegraphics[width=\linewidth]{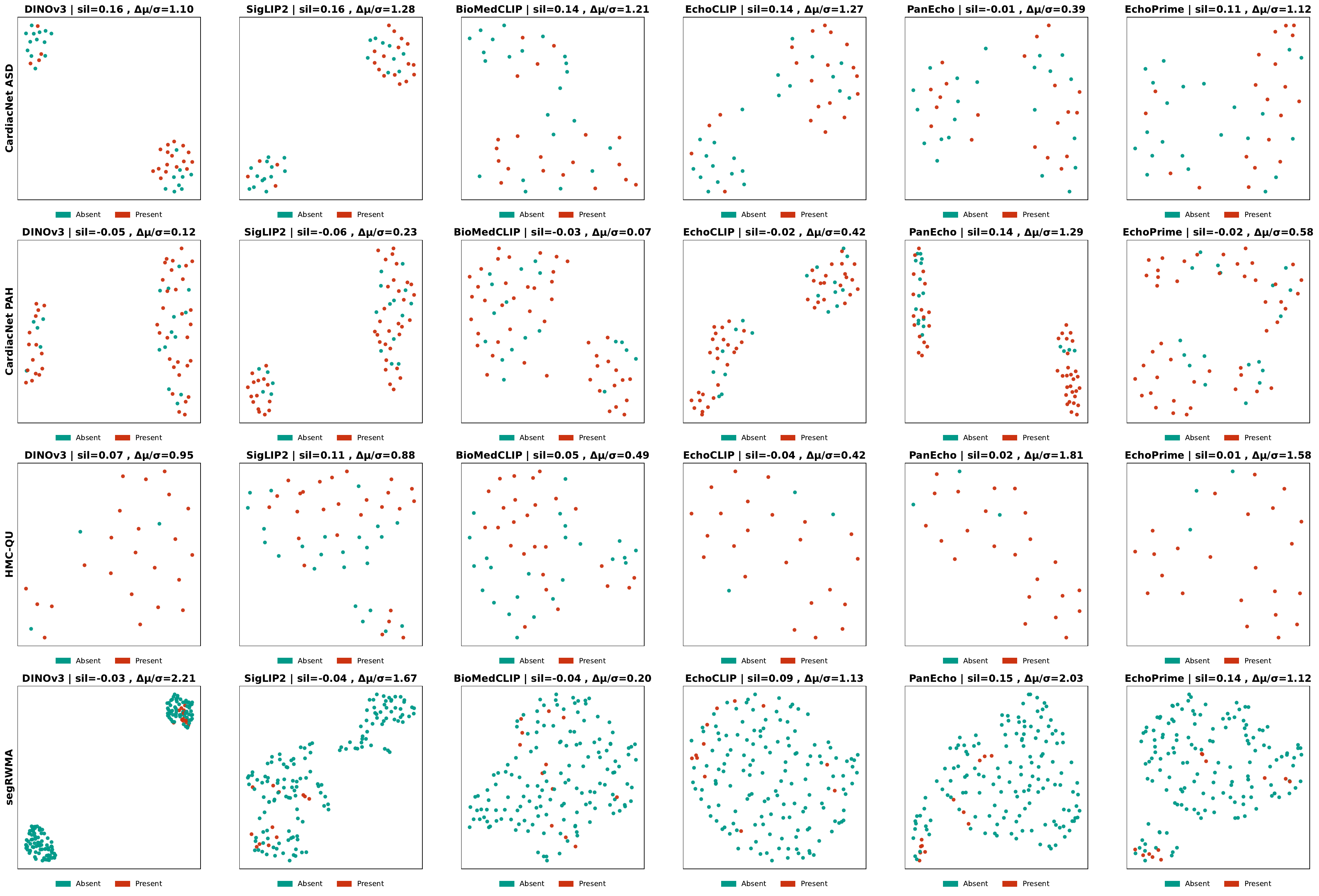}
    \caption{UMAP of visual representations on CardiacNet, HMC-QU, and segRWMA datasets. These projections visualize the clustering behavior of embeddings for pathology classification tasks.}
    \label{fig:vis_class}
\end{figure}

\textbf{Embedding structures for pathology tasks.} Within CardioBench, inspection of the embedding spaces for classification tasks evidences that zero-shot performance is constrained by weakly discriminative representation spaces.
The UMAP visualizations in Figure \ref{fig:vis_class}, pathology-present and pathology-absent cases form partially separable but substantially overlapping clusters, with limited intra-class compactness and low silhouette scores across datasets. This indicates the limited prioritization of pathology-specific cues in current visual backbones, which tend instead to encode broader distributional features. The contrast with linear probing, showing substantially higher performance for BioMedCLIP and SigLIP2, further highlights that discriminative signals are present but not aligned with text prompts or directly accessible for zero-shot. These findings underscore the gap between latent signal and usable representation, emphasizing the need for models that organize clinical information more explicitly.

CardioBench makes clear that progress in echocardiography foundation models cannot be measured by zero-shot performance alone. Across regression, classification, and view recognition, the benchmark reveals a consistent pattern: models contain latent clinical signal, but its accessibility depends heavily on architectural design, training supervision, and the stability of the embedding organization. This points to several practical directions. First, explicit supervision for core clinical axes such as EF or view classification proves more reliable than expecting them to emerge implicitly, suggesting that pretraining pipelines should integrate lightweight but structured supervision. Second, temporal modeling is indispensable for functional tasks, as demonstrated by PanEcho, while retrieval-based matching offers complementary robustness, motivating hybrid approaches that combine the strengths of both. Third, domain-specific text encoders, as in EchoCLIP, can enforce physiologically meaningful cross-modal structure, but their advantage is not stable, underscoring the need to broaden cardiac text corpora. Finally, the surprisingly strong performance of general-purpose encoders such as SigLIP2 and DINOv3 highlights both an opportunity and a limitation: scale and diversity alone can produce robust baselines under domain shift, yet these models fail to organize clinical signals in a way that supports fine-grained reasoning. This suggests that future cardiac foundation models should not discard generalist architectures, but rather adapt them through targeted supervision and domain grounding, bridging the gap between robustness and clinical fidelity.

\section{Conclusion}
CardioBench demonstrates that echocardiography foundation models must be assessed through multi-evaluation to capture their true capabilities. Performance depends on design and supervision choices shaping temporal dynamics, retrieval, and clinically grounded representations. Future advances will likely come from hybrid approaches combining these strengths. By providing a publicly available standardized benchmark, CardioBench establishes a baseline for fair comparison to develop clinically meaningful models.

% \section{Reproducibility Statement}  
% Details are provided in Supplementary Material, Section~\ref{app:reproducibility}, and all code and resources are available at \url{https://anonymous.4open.science/r/CardioBench/}.  

\clearpage
\bibliographystyle{named}
\bibliography{ijcai26}
\clearpage
\appendix
\section{Models}
\label{app:models}
\begin{table}[b]
\centering
\scriptsize
% Reduce horizontal padding to the absolute minimum
\setlength{\tabcolsep}{2pt} 
\renewcommand{\arraystretch}{1.3}

\begin{tabularx}{\columnwidth}{l|c|c|c|X} % X column will wrap text
\hline
\textbf{Model} & \textbf{Vision} & \textbf{Text} & \textbf{Temp.} & \textbf{Training Data}\\
\Xhline{1pt}
\textbf{EchoCLIP}   & ConvNeXt-B & CLIP & – & 1.03M A4C videos + reports \\
\textbf{EchoPrime}  & mViT       & B-BERT & Video & 12.1M MV videos + reports \\
\textbf{PanEcho}    & CNeXt-T    & –      & Trans. & 1.2M MV echo videos \\
\textbf{EchoFM}     & ViT-L/16   & –      & Video & 290K MV echo videos \\
\textbf{BioMedCLIP} & ViT-B/16   & P-BERT & – & 15M image-caption pairs \\
\textbf{DINOv3}     & ViT-L/16   & –      & – & 1.7B natural images \\
\textbf{SigLIP2}    & ViT-B/16   & ViT    & – & 10B WebLI images \\
\hline
\end{tabularx}
\caption{Summary of evaluated foundation models.}
\label{tab:ModelOverview}
\end{table}

Table~\ref{tab:ModelOverview} provides a high-level comparison, while below each model is described in more detail. CardioBench compares echocardiography-specific, biomedical, and general-purpose foundation models. \textbf{EchoCLIP} \cite{christensen2024vision} adapts a ConvNeXt-B vision encoder with a CLIP-style text tower, trained contrastively on 1M A4C echo videos and reports, aligning video embeddings with task-specific prompts at inference. \textbf{EchoPrime} \cite{vukadinovic2024echoprime} combines a multiview ViT (mViT) with BioMedBERT and uses retrieval, projecting test videos into a joint embedding space and predicting by matching to labeled exemplars. \textbf{PanEcho} \cite{holste2025panecho} employs a ConvNeXt-T backbone with a temporal frame transformer, trained on 1.2M multiview echo videos for multitask regression and classification. \textbf{EchoFM} \cite{kim2024echofm} uses a ViT-L/16 video encoder trained on 290K multiview echo videos to learn general embeddings optimized for probing. As the linear heads are not provided, and the model doesn't have the text encoder, zero-shot cannot be performed \textbf{BioMedCLIP} \cite{zhang2023BioMedCLIP} pairs a ViT-B/16 with PubMedBERT, pretrained on 15M biomedical image–text pairs spanning radiology, pathology, microscopy, and ultrasound. \textbf{DINOv3} \cite{simeoni2025dinov3} is a self-supervised ViT-L/16 trained on 1.7B natural images with an aligned text encoder, applied by encoding frames and pooling temporally before probing or computing similarity with handcrafted prompts. Finally, \textbf{SigLIP2} \cite{tschannen2025siglip} is a multilingual vision–language model with a ViT-B/16 backbone and transformer text tower, trained on 10B WebLI pairs. Together, these models allow us to assess how far both biomedical and large-scale generic supervision can be transferred to echocardiography tasks, and whether modality-specific pretraining is necessary to achieve competitive performance.

\section{Datasets}
\label{app:datasets}

In this section, we motivate the choice of datasets for evaluation, provide the distribution of values in each dataset, and describe the splitting strategy.

\subsection{Dataset Selection}
Because echocardiography involves sensitive patient information, the number and size of public datasets are limited. We use eight datasets that are either openly downloadable or available upon request. Table~\ref{tab:datasets} summarizes their key characteristics.

\begin{table}[t]
\centering
\scriptsize
\setlength{\tabcolsep}{3pt} 
\renewcommand{\arraystretch}{1.2}

\begin{tabularx}{\columnwidth}{l|X|l|l}
\hline
\textbf{Dataset} & \textbf{Source} & \textbf{Availability} & \textbf{Type} \\
\hline
\textbf{EchoNet-Dyn.} \cite{echonetdynamic2019}  & Stanford AIMI & Open & Video \\
\textbf{EchoNet-Ped.} \cite{echonetpediatric2025} & Stanford AIMI & Open & Video \\
\textbf{EchoNet-LVH} \cite{echonetlvh2020}       & Stanford AIMI & Open & Video \\
\textbf{SegRWMA} \cite{liu2023enhance}           & Kaggle        & Open & Video \\
\textbf{CardiacNet} \cite{yang2024cardiacnet}    & Kaggle        & Open & Video \\
\textbf{CAMUS} \cite{leclerc2019deep}            & Univ. de Lyon & Open & Video \\
\textbf{HMC-QU} \cite{degerli2021early}          & Private       & Request & Video \\
\textbf{TMED-2} \cite{huang2022tmed}             & Private       & Request & Image \\
\hline
\end{tabularx}
\caption{Echocardiography datasets used in this study, with their source, accessibility, and modality.}
\label{tab:datasets}
\end{table}

\begin{table}[t]
\centering
\scriptsize
\setlength{\tabcolsep}{2pt}
\renewcommand{\arraystretch}{1.1}
\begin{tabularx}{\columnwidth}{l|c|c|X|X}
\hline
\textbf{Dataset} & \textbf{Size} & \textbf{T/V/T} & \textbf{Labels Used} & \textbf{View} \\
\hline
EchoNet-Dyn. \greencircle  & 10,030 v & 7465/1288/1277 & EF & A4C \\
EchoNet-Ped. \greencircle & 7,810 v  & 6365/798/658    & Age, Sex, Weight, Height & A4C \\
EchoNet-LVH \greencircle      & 12,000 v & 10.5k/1.2k/343 & IVSd, LVIDd, LVPWd & PLAX \\
SegRWMA           & 529 v   & 221/152/156          & RWMA & A4C, A3C, A2C \\
CardiacNet-ASD      & 228 v   & 158/23/47          & ASD & A4C \\
CardiacNet-PAH      & 471 v   & 319/51/106          & PAH & A4C \\
CAMUS \greencircle & 1,000 v  & 400/50/50         & EF, Sex, Age, Quality & A4C, A2C \\
HMC-QU            & 322 v   & 227/45/50          & STEMI & A4C, A2C \\
TMED-2 \greencircle  & 17,270 i   & 360/119/119  & AS & A4C, A2C, PSAX, PLAX \\
\hline
\end{tabularx}
\caption{Summary of dataset characteristics. 'v' denotes videos, 'i' images. \greencircle indicates official splits.}
\label{tab:dataset_stats}
\end{table}

Table~\ref{tab:dataset_stats} provides an overview of dataset sizes, experimental splits, and available labels. For the CAMUS and TMED-2 datasets, we report the total number of unique videos and images, with splits defined at the patient level. For the other datasets, we assume one video per patient. We also indicate the type of annotations provided and describe how the data were partitioned into training, validation, and testing sets. Where applicable, we additionally summarize the distribution of classes.

\subsection{Dataset Details}
\begin{figure}[t]
    \centering
    \begin{subfigure}[t]{0.32\textwidth}
        \centering
        \includegraphics[width=\linewidth]{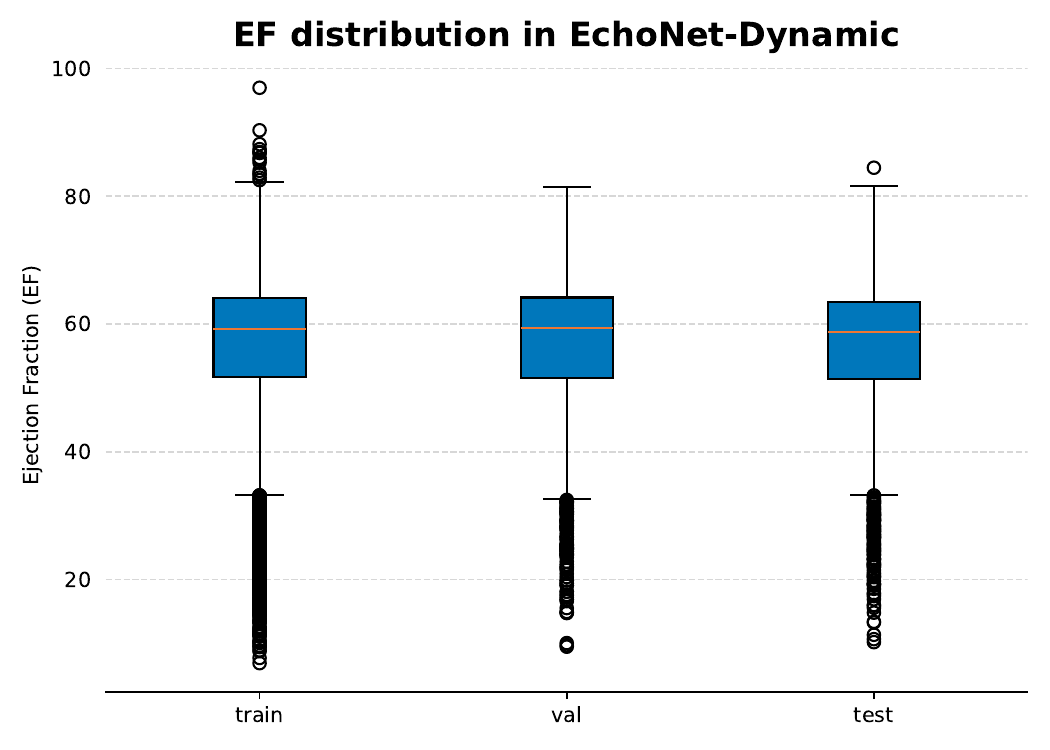}
        \caption{EchoNet-Dynamic EF}
        \label{fig:echonetdynamic_ef}
    \end{subfigure}
    \hfill
    \begin{subfigure}[t]{0.32\textwidth}
        \centering
        \includegraphics[width=\linewidth]{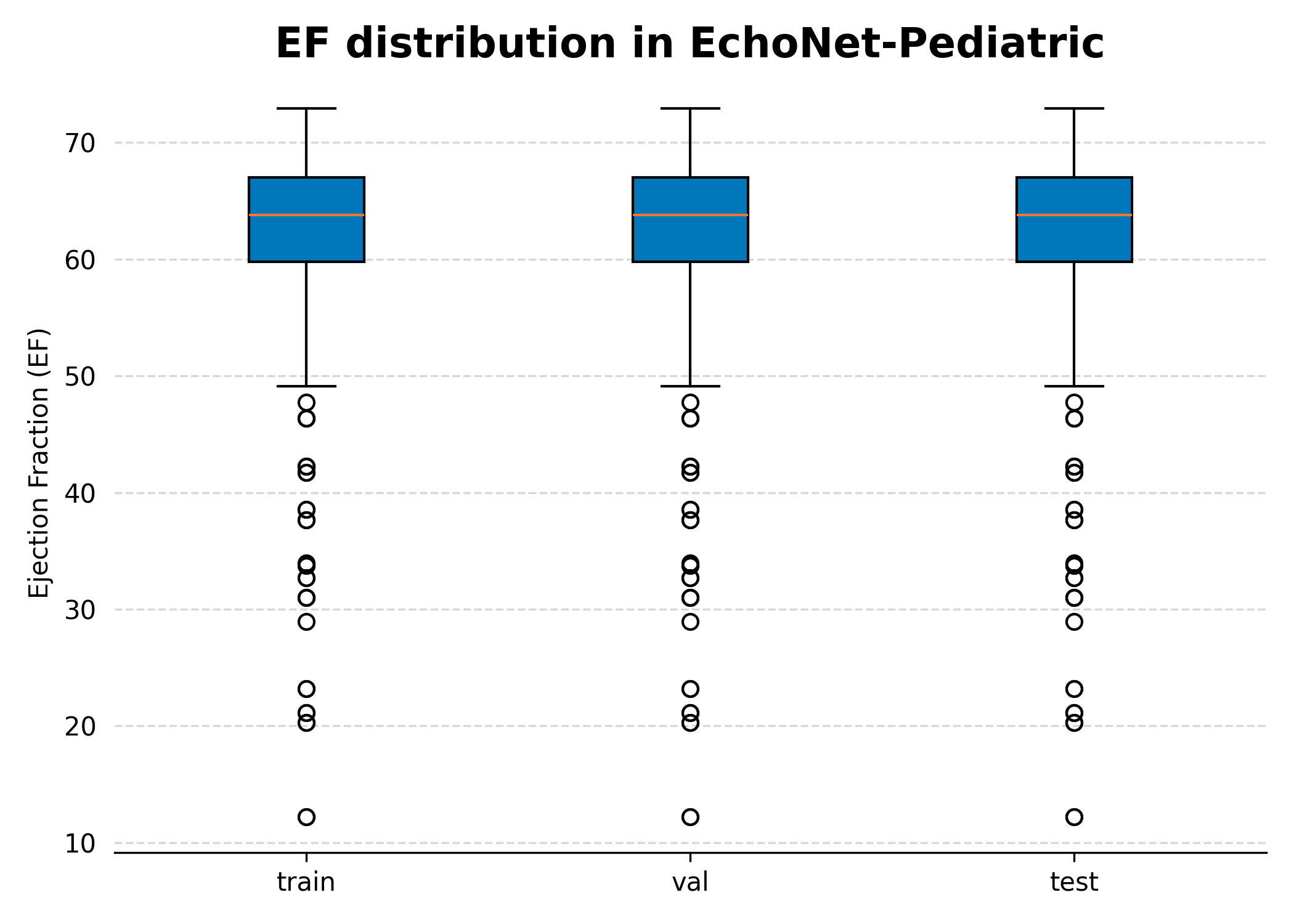}
        \caption{EchoNet-Pediatric EF}
        \label{fig:echonetped_ef}
    \end{subfigure}
    \hfill
    \begin{subfigure}[t]{0.32\textwidth}
        \centering
        \includegraphics[width=\linewidth]{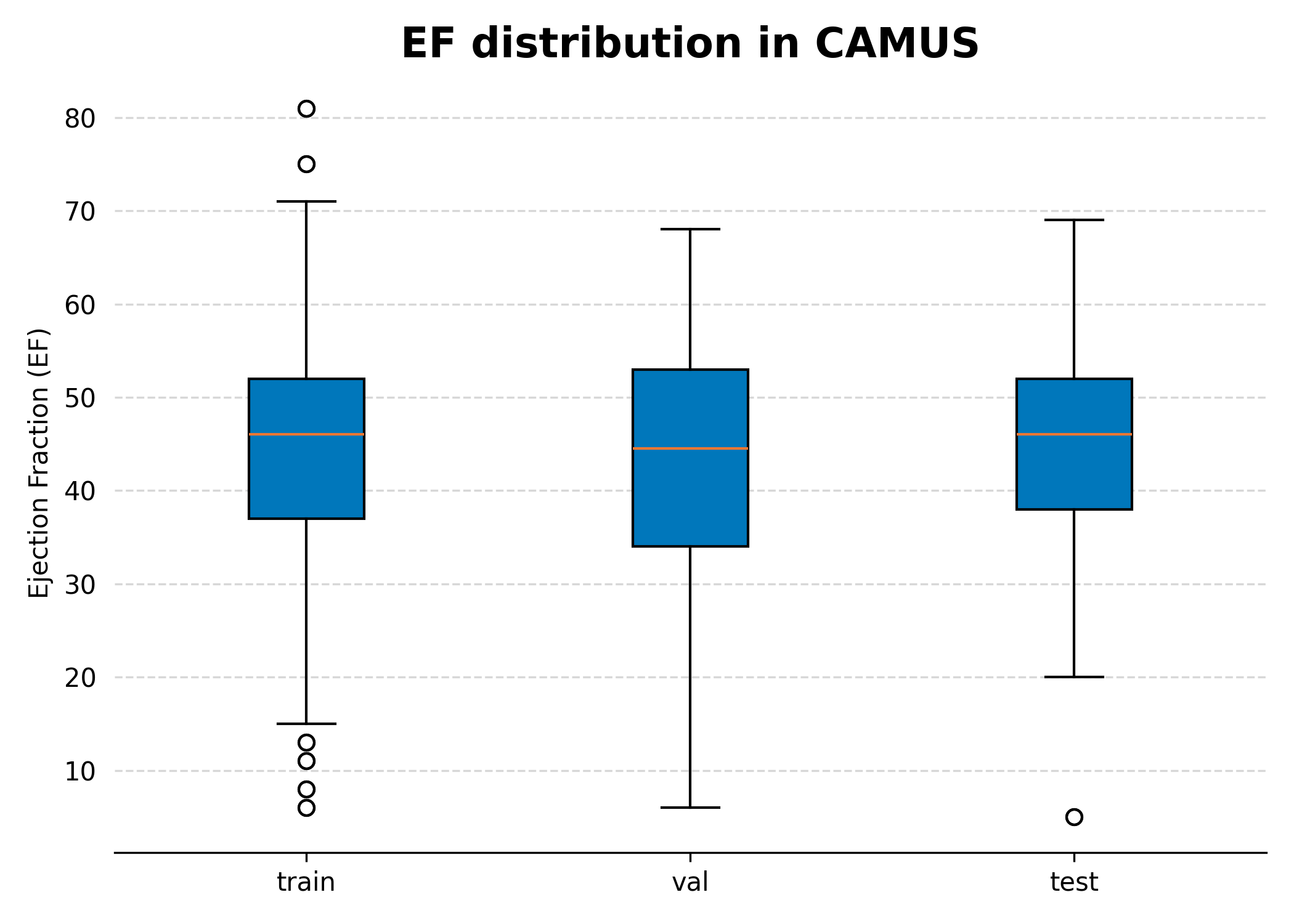}
        \caption{CAMUS EF}
        \label{fig:camus_ef}
    \end{subfigure}
    \caption{Box plots of EF distributions across three datasets: EchoNet-Dynamic, EchoNet-Pediatric, and CAMUS.}
    \label{fig:case_distributions}
\end{figure}

\paragraph{EchoNet-Dynamic.} 
The dataset consists of 10,030 A4C echocardiography videos, each from a unique patient. Every video is annotated with an EF value, with the distribution shown in Figure~\ref{fig:echonetdynamic_ef}.

\begin{figure}[t]
    \centering
        \centering
        \includegraphics[width=\linewidth]{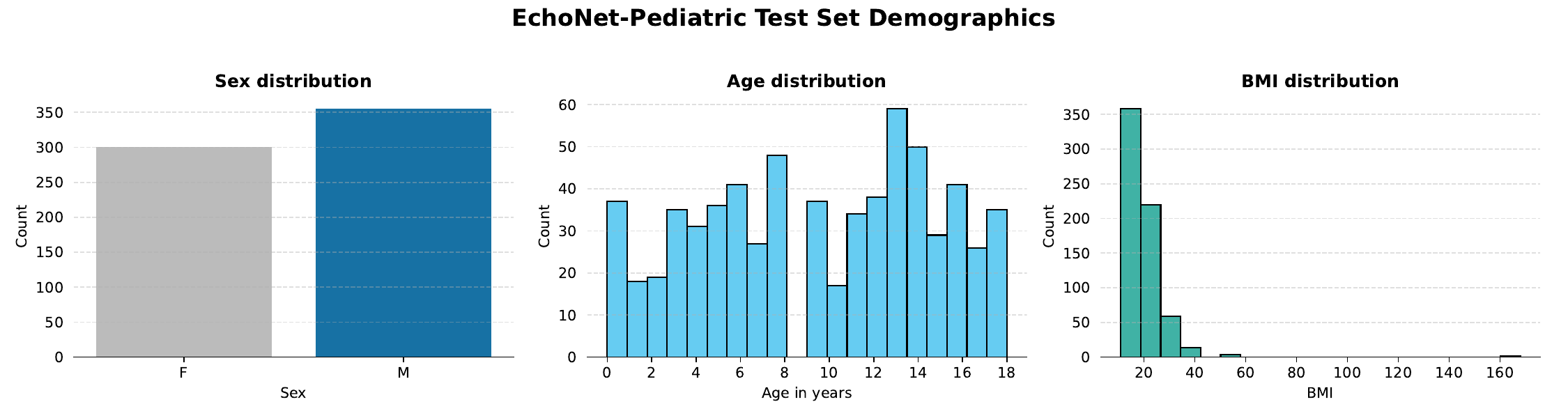}
    \caption{Distribution of sex, age, and BMI for video samples in the EchoNet-Pediatric dataset.}
    \label{fig:ped_demo_distr}
\end{figure}

\begin{figure}[t]
    \centering
        \centering
        \includegraphics[width=\linewidth]{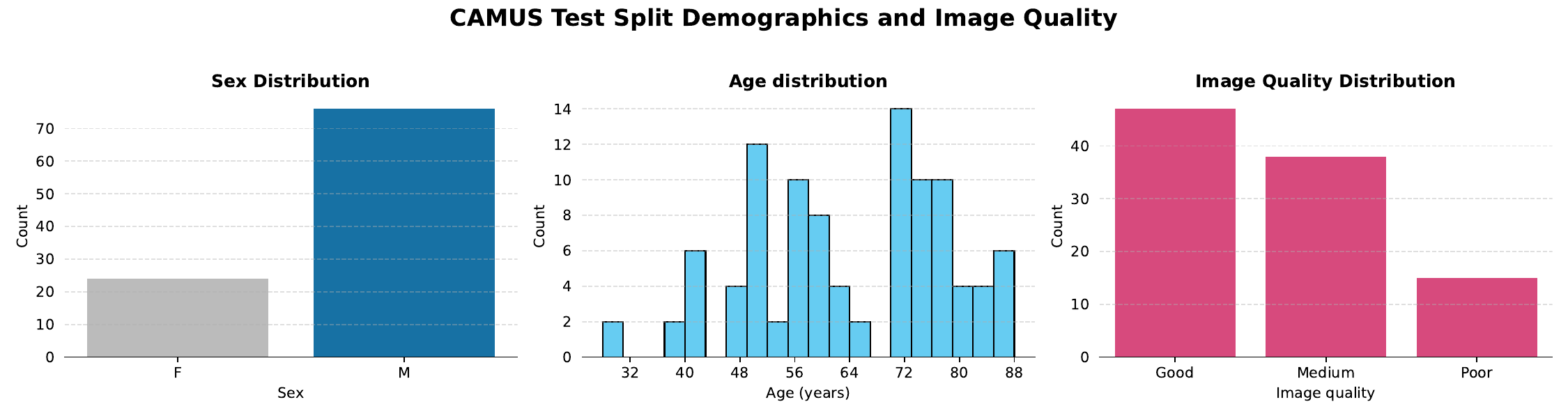}
    \caption{Distribution of sex, age, and image quality in the CAMUS dataset.}
    \label{fig:camus_demo_distr}
\end{figure}

\begin{figure}[t]
    \centering
    \includegraphics[width=\linewidth]{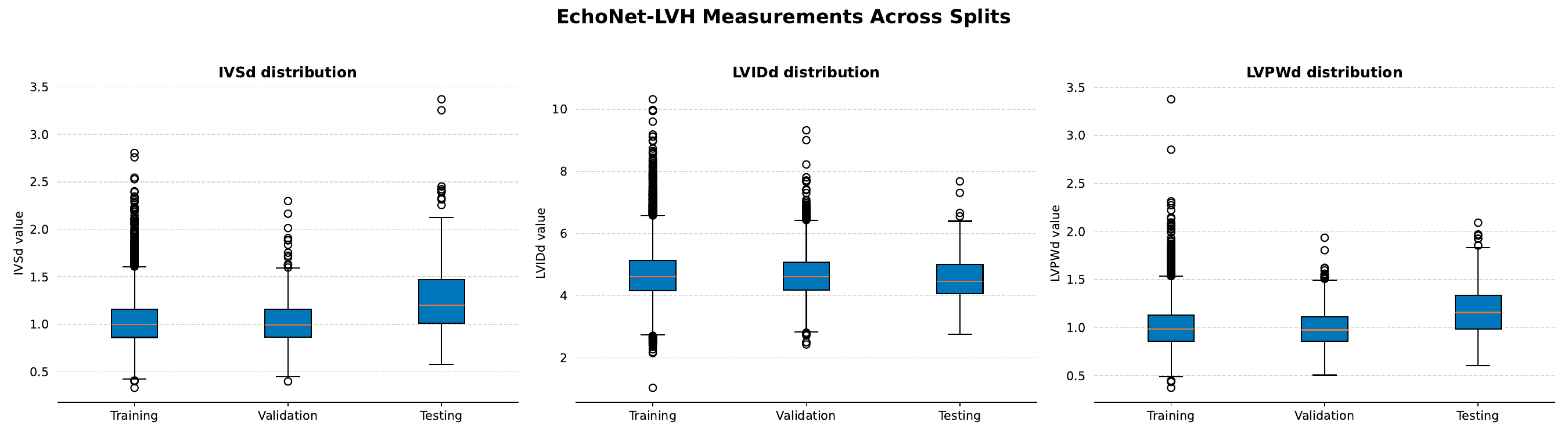}
    \caption{Distribution of structural measurements in the EchoNet-LVH dataset.}
    \label{fig:lvh_distr}
\end{figure}

\paragraph{EchoNet-Pediatric.} 
The dataset comprises 7,810 videos, including 4,526 PSAX and 3,284 A4C echocardiography recordings, with one video per patient. Each video is annotated with EF, sex, age, weight, and height, from which body mass index (BMI) is derived. The EF distribution is shown in Figure~\ref{fig:echonetped_ef}, and the demographic distributions are presented in Figure~\ref{fig:ped_demo_distr}.

\paragraph{EchoNet-LVH.}
The EchoNet-LVH dataset contains 12,000 PLAX-view videos, each annotated with the frame on which structural measurements (IVSd, LVIDd, LVPWd) are performed, with their distributions shown in Figure~\ref{fig:lvh_distr}.

\paragraph{CAMUS.}
The CAMUS dataset comprises 500 patients, each with two echocardiography views (A2C and A4C). Each video is annotated with sex, age, EF, and image quality. We follow the official split of 400 patients for training, 50 for validation, and 50 for testing. The EF distribution is shown in Figure~\ref{fig:camus_ef}, and the demographic distributions are presented in Figure~\ref{fig:camus_demo_distr}.

\begin{figure}[t]
    \centering
    \includegraphics[width=\linewidth]{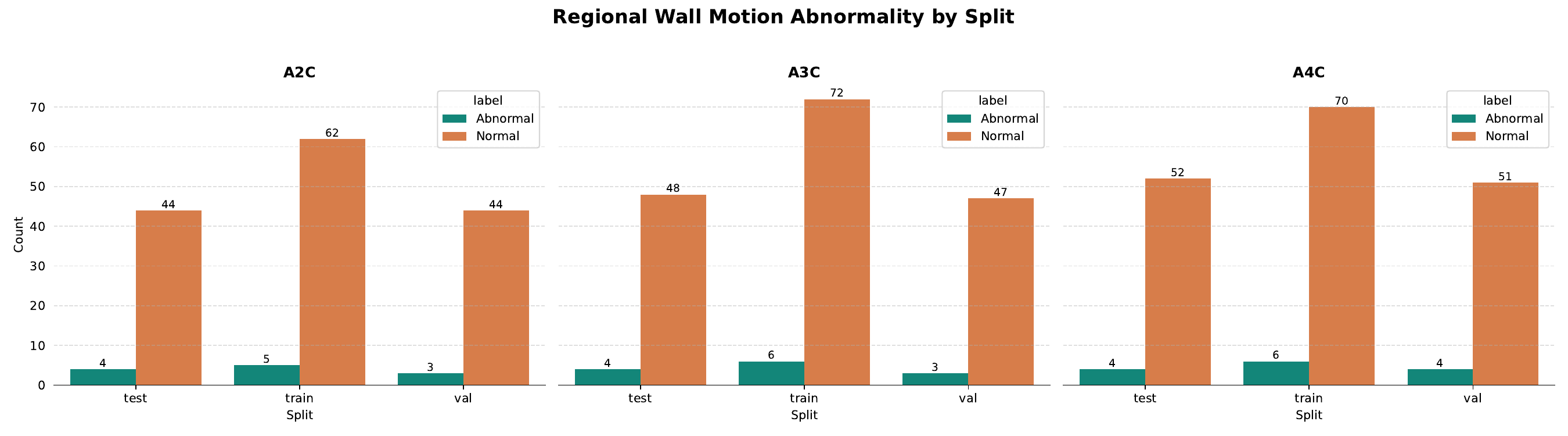}
    \caption{Distribution of regional wall motion abnormalities in the SegRWMA dataset across A2C, A3C, and A4C views and dataset splits.}
    \label{fig:rwma_split}
\end{figure}

\paragraph{SegRWMA.}
The SegRWMA dataset includes 198 patients with regional wall motion annotations, comprising 14 abnormal cases in the A4C view, 13 in the A3C view, and 12 in the A2C view, with the remaining patients considered normal. Segmentation masks are provided for the annotated frames, and we use the first annotated frame index for evaluation. In this study, we restrict analysis to the 2D ultrasound modality, as it is more cost-effective than contrast-enhanced echocardiography~\cite{liu2023enhance}. To prevent data leakage, the dataset is split at the patient level, ensuring that no patient appears in multiple splits. As shown in Figure~\ref{fig:rwma_split}, the abnormality distribution is imbalanced across splits: in the A2C view, 4 abnormal patients are in training, 5 in testing, and 3 in validation; in the A3C view, 6 are in training, 4 in testing, and 3 in validation; and in the A4C view, 6 are in training, 4 in testing, and 4 in validation. The remaining patients in each split are normal.

\begin{figure}[t]
    \centering
    \includegraphics[width=\linewidth]{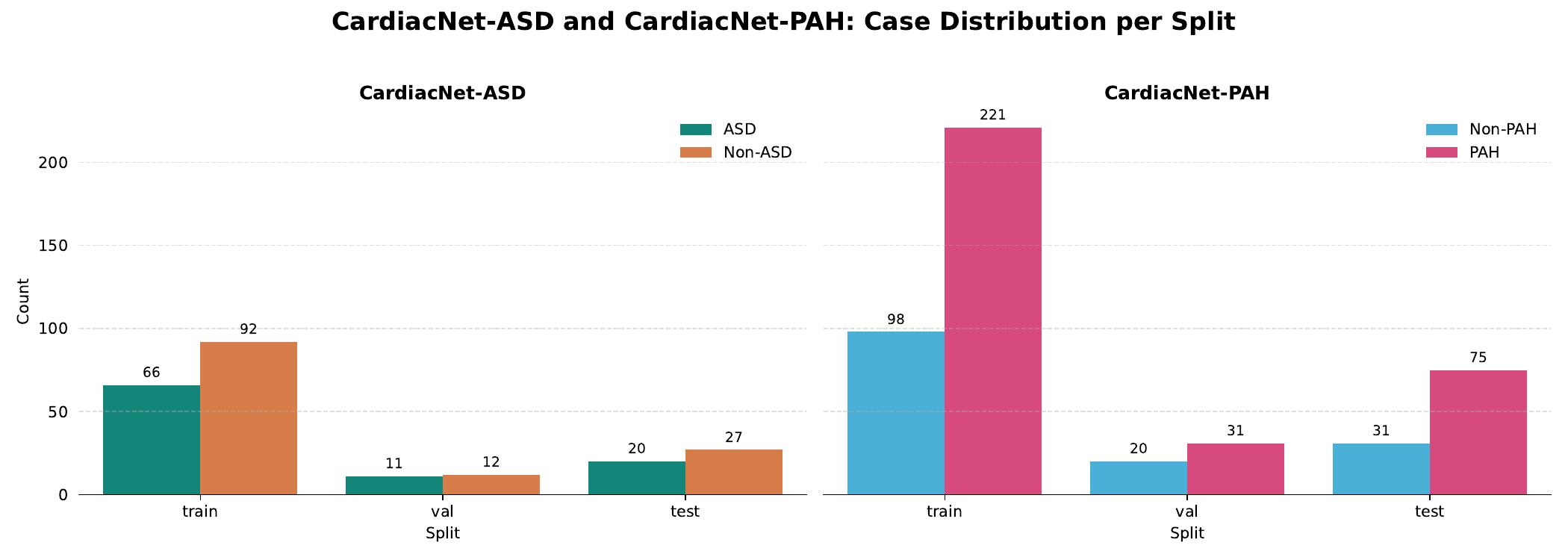}
    \caption{Distribution of binary labels in the CardiacNet dataset for ASD and PAH across training, validation, and test splits.}
    \label{fig:cardiacnet_split}
\end{figure}

\paragraph{CardiacNet.}
The CardiacNet dataset contains 228 videos for ASD and 529 videos for PAH. Following the authors~\cite{yang2024cardiacnet}, we treat each video as a separate patient. The dataset is divided independently for each task according to its distribution. For the CardiacNet-ASD subset, we apply a stratified split to preserve the proportion of ASD and non-ASD cases across subsets: 20\% of patients are held out for testing, while the remaining 80\% are further split, with 12.5\% allocated to validation. For the CardiacNet-PAH subset, we use patient-level labels and again perform a stratified split to preserve the proportion of PAH and non-PAH cases: 20\% of patients are reserved for testing, and from the remaining 80\%, 12.5\% are allocated to validation. The distribution of binary labels across splits for both ASD and PAH tasks is shown in Figure~\ref{fig:cardiacnet_split}.

\begin{figure}[t]
    \centering
    \begin{subfigure}[b]{0.48\linewidth}
        \centering
        \includegraphics[width=\linewidth]{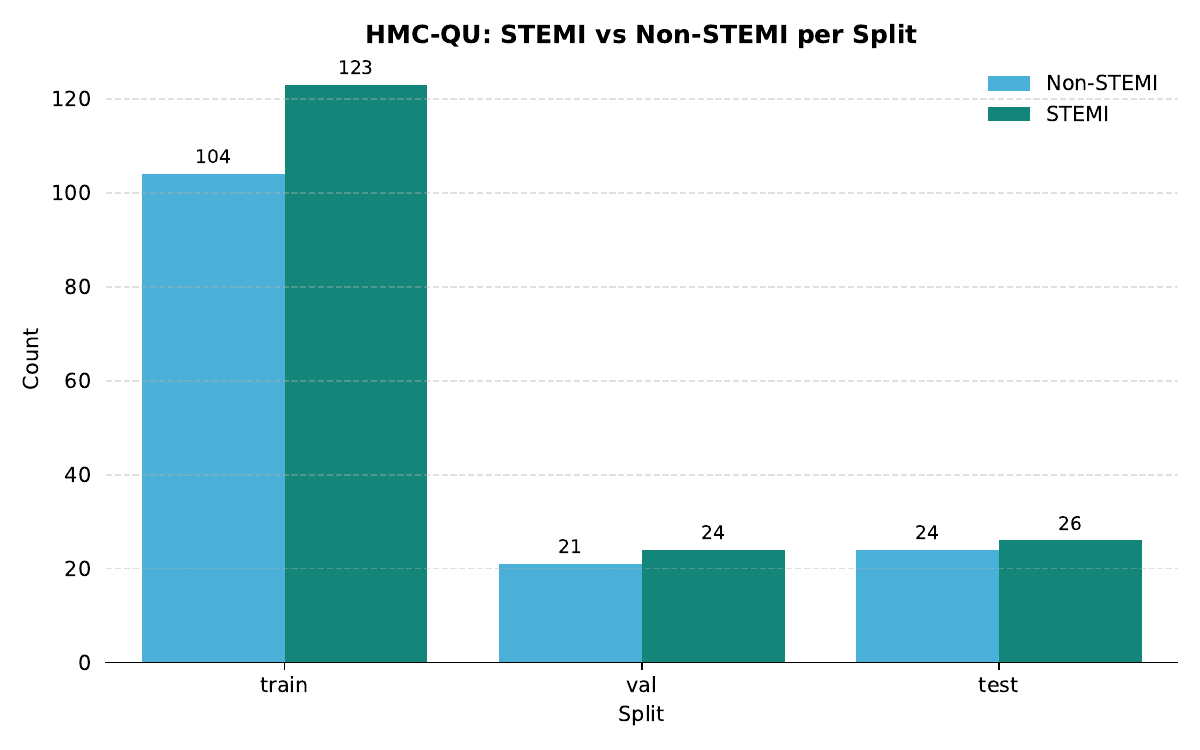}
        \caption{HMC-QU}
        \label{fig:hmcqu_split}
    \end{subfigure}
    \hfill
    \begin{subfigure}[b]{0.48\linewidth}
        \centering
        \includegraphics[width=\linewidth]{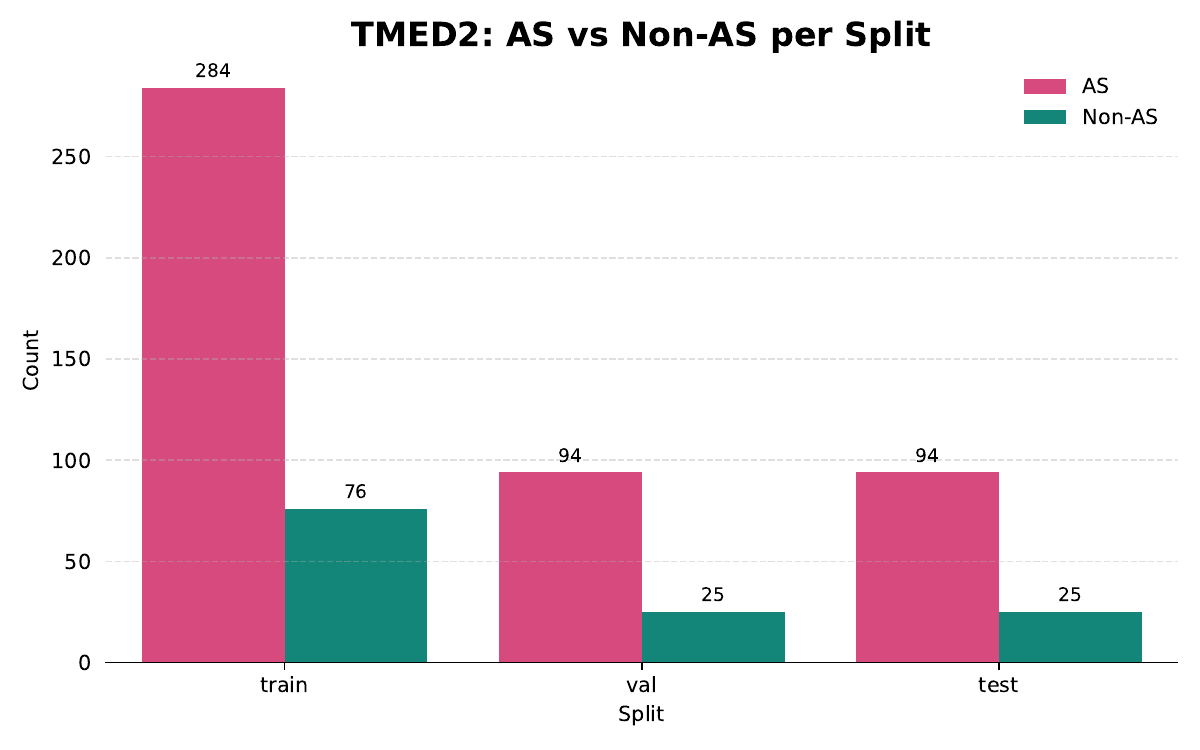}
        \caption{TMED-2}
        \label{fig:tmed2_split}
    \end{subfigure}
    \caption{Overview of label distributions across splits for the HMC-QU and TMED-2 datasets.}
    \label{fig:split_distributions}
\end{figure}

\paragraph{HMC-QU.}
The HMC-QU dataset contains 332 videos of A4C and A2C views with STEMI labels. Using patient-level labels, we apply a stratified split to maintain the STEMI/non-STEMI ratio across subsets. The dataset is divided into approximately 70.8\% for training, 14.2\% for validation, and 15\% for testing, ensuring that all videos from the same patient remain in a single subset. We treat each video as a separate test case due to the relatively small dataset size. The distribution of STEMI and non-STEMI cases across splits is shown in Figure~\ref{fig:hmcqu_split}.

\paragraph{TMED-2.}
TMED-2 is the only image dataset in our study, comprising 17,270 images across views: 1,670 A2C, 2,206 A4C, 4,808 PLAX, 1,725 PSAX, and 6,861 labeled as Other (A2C, A4C, or other views). Since many images belong to the same study, they are grouped into 598 studies in total. Following the official DEV479 split, the dataset is partitioned into 360 studies for training, 119 for validation, and 119 for testing. We also binarize the labels from multiclass classification into aortic stenosis "present" and "absent." The distribution of binary aortic stenosis labels across splits is presented in Figure~\ref{fig:tmed2_split}.

\section{Reproducibility}
\label{app:reproducibility}

Each ultrasound video $V \in \Re^{T \times H \times W }$ is represented by 16 consecutive frames, normalized and resized to $224 \times 224$, yielding $X \in \Re^{16 \times 224 \times 224}$. A video encoder $f_\theta$ produces an embedding $z_{v}=f_\theta(X) \in \Re^{d}$, while a text prompt $P$ is mapped by a text encoder $g_\theta$ into $z_{p}=g_\theta(P) \in \Re^{d}$. For models originally designed for single images, we extend them to videos by computing predictions frame-wise and reporting the mean of the outputs across the 16 frames.  

\textbf{Zero-shot evaluation.} For classification, we define one prompt per class ($P_{1}, \dots, P_{k}$) and predict using cosine similarity: $\hat{y} = \arg\max_{c} \cos(z_{v}, z_{p_c}).$  
This $\arg\max$ rule avoids dataset-specific thresholds, ensuring a calibration-free and reproducible evaluation. For regression tasks, we follow \cite{christensen2024vision} by constructing prompts with numerical values over a predefined range. Predictions are obtained by aggregating frame-wise similarities (median of the top 20\% per frame, averaged across frames). Prompt templates are detailed in Section~\ref{app:prompts}.

\textbf{Probing.} We assess the quality of the learned representations by applying two lightweight classifiers directly on the embedding space. First, we perform linear probing by freezing the model’s parameters and training a linear classifier on top of the embeddings. Linear probing tests whether the information needed for a task is linearly accessible. Second, for the view classification task, we apply $k$-nearest neighbor (kNN) classification directly in the embedding space. Unlike linear probing, kNN evaluates whether local structure in the embedding space naturally reflects clinically meaningful view categories. By combining linear probing for global linear separability with kNN for local structure, we obtain complementary insights into how foundation models encode clinical information.

Training is carried out using the AdamW optimizer on the linear head only using a learning rate of 1e-4 with a weight decay of 1e-2. We use a batch size of 64, applying cross-entropy loss for classification tasks and mean squared error (MSE) loss for regression tasks. Early stopping is applied on the validation split to prevent overfitting. All experiments are conducted on an NVIDIA RTX A6000 GPU.

\section{Prompts}
\label{app:prompts}

\begin{table}[h]
\centering
\tiny 
\setlength{\tabcolsep}{2pt}
\renewcommand{\arraystretch}{1.2}

\begin{tabularx}{\columnwidth}{l X}
\toprule
\textbf{Task} & \textbf{Prompt Templates} \\
\midrule
\texttt{EF} & \ttfamily "THE LEFT VENTRICULAR EJECTION FRACTION IS ESTIMATED TO BE <\#>\%", "LV EJECTION FRACTION IS <\#>\%." \\
\midrule
\texttt{LVIDd} & \ttfamily "LEFT VENTRICULAR INTERNAL DIAMETER IN DIASTOLE (LVIDD) IS <\#> CM.", "LVIDD IS <\#> CM." \\
\midrule
\texttt{AS (+)} & \ttfamily "AORTIC STENOSIS IS PRESENT.", "SEVERE AORTIC STENOSIS.", "CALCIFIED AORTIC VALVE WITH RESTRICTED LEAFLET MOTION." \\
\midrule
\texttt{AS (-)} & \ttfamily "NO AORTIC STENOSIS.", "NO SIGNIFICANT AORTIC VALVE STENOSIS.", "AORTIC VALVE OPENS NORMALLY WITHOUT STENOSIS." \\
\bottomrule
\end{tabularx}
\caption{Condensed format of prompt templates. Regression prompts use numerical placeholders $<\#>$ replaced during evaluation.}
\label{tab:prompts_tiny}
\end{table}
The prompt design follows the standard established by \cite{christensen2024vision}. Their exact ejection fraction prompt is used directly, while the prompts for the remaining tasks are generated in accordance with the same style. To improve robustness and reduce prompt-specific bias, we instantiate multiple phrasings per class (classification) or per numeric value (regression). For classification, the mean similarity is computed separately for each class and the class with the higher mean is selected. For regression, numerical placeholders are replaced with candidate values from a predefined grid, and the value corresponding to the prompt with the highest similarity is selected as the prediction. Specifically, ejection fraction is instantiated over integer values from 0–100\%, while chamber dimensions and wall thicknesses are instantiated over clinically reasonable ranges with 0.1\,cm resolution: LVIDd from 2.0–8.0\,cm, IVSd from 0.5–2.0\,cm, and LVPWd from 0.5–2.0\,cm. All prompts and ranges are released on GitHub to ensure reproducibility.

\end{document}